\documentclass[11pt,a4paper]{article}
\usepackage[hyperref]{emnlp2020}

\usepackage{times}
\usepackage{latexsym}

\usepackage{graphicx}
\usepackage{amsmath}
\usepackage{algorithmic}
\usepackage{algorithm}

\usepackage{microtype}
\usepackage{color}

\aclfinalcopy

\title{Efficient Transformer-based Large Scale Language Representations using Hardware-friendly Block Structured Pruning}

\author{
    Bingbing Li \thanks{These authors contributed equally}  \textsuperscript{\rm 1},
    Zhenglun Kong  \footnotemark[1]  \textsuperscript{\rm 2},
    Tianyun Zhang \textsuperscript{\rm 3},
    Ji Li \textsuperscript{\rm 4},
    Zhengang Li \textsuperscript{\rm 2},
    Hang Liu \textsuperscript{\rm 5},
    Caiwen Ding \textsuperscript{\rm 1}\\
 $^1$University of Connecticut,
 $^2$Northeastern University,
 $^3$Syracuse University,\\
 $^4$Microsoft Corporation,
 $^5$Stevens Institute of Technology\\
 \{bingbing.li, caiwen.ding\}@uconn.edu,
 \{kong.zhe, li.zhen\}@northeastern.edu,
 tzhan120@syr.edu,\\
 changzhouliji@gmail.com,
 hliu77@stevens.edu
 }

\date{}

\begin{document}
\maketitle
\begin{abstract}
Pre-trained large-scale language models have increasingly demonstrated high accuracy on many natural language processing (NLP) tasks. However, the limited weight storage and computational speed on hardware platforms have impeded the popularity of pre-trained models, especially in the era of edge computing. In this work, we propose an efficient transformer-based large-scale language representation using hardware-friendly block structure pruning. We incorporate the reweighted group Lasso into block-structured pruning for optimization.
Besides the significantly reduced weight storage and computation, the proposed approach achieves high compression rates. Experimental results on different models (BERT, RoBERTa, and DistilBERT) on the General Language Understanding Evaluation (GLUE) benchmark tasks show that we achieve up to 5.0$\times$ with zero or minor accuracy degradation on certain task(s). Our proposed method is also orthogonal to existing compact pre-trained language models such as DistilBERT using knowledge distillation, since a further 1.79$\times$ average compression rate can be achieved on top of DistilBERT with zero or minor accuracy degradation. It is suitable to deploy the final compressed model on resource-constrained edge devices.
\end{abstract}

\section{Introduction}

Transformer-based language model pre-training has proven to be highly effective in learning universal language representations from large-scale unlabeled data and being fine-tuned to adapt to downstream tasks \cite{peters2018deep,sun2019patient}.
Representative works such as BERT~\cite{devlin2018bert}, XLNet~\cite{yang2019xlnet}, RoBERTa~\cite{liu2019roberta}, MT-DNN~\cite{liu2019multi},  ALBERT~\cite{lan2019albert}, GPT-2~\cite{radford2019language}, and UniLMv2~\cite{bao2020unilmv2} have substantially advanced the state-of-the-art across a variety of downstream tasks, such as text classification, natural language inference, and question answering. 

Despite its success in performance improvement in natural language understanding and generation, the computational cost and data storage of Transformer-based pre-trained language model are two widely recognized concerns due to Transformer's deep architecture and rich parameters. These models typically contain several hundred million parameters. The recent released research models even reach multi-billion parameters, such as MegatronLM (8.3 billion parameters)~\cite{shoeybi2019megatron}, Turing-NLG (17 billion parameters)~\cite{turing} and GPT-3 (175 billion parameters)~\cite{gpt3}, which require more advanced computing facility.
Hence, it is imperative to reduce the computational cost and model storage of pre-trained Transformer-based language models in order to popularize their applications in computer systems, especially in edge devices with limited resources. 

Several works have been developed in the context of model compression, such as knowledge distillation~\cite{hinton2015distilling,sanh2019distilbert,jiao2019tinybert,sun2019patient}, weight pruning~\cite{han2015learning}, parameter sharing~\cite{lan2019albert} and weight quantization~\cite{polino2018model}. For computer vision, the information compressed/reduced in image features can be partially retrieved from neighboring pixels since they share similar and uniform characteristics spatially. However, for NLP, the syntax and semantics information of Transformer in language/text domain are more sensitive than that of computer vision. A high compression rate for large-scale language models is difficult to achieve on downstream NLP tasks. As a result, there are few works in exploring and optimizing hardware-friendly model compression techniques for state-of-the-art Transformer-based pre-trained language models, to reduce the weight storage and computation on computer system while maintaining prediction accuracy. 

In this work, we propose an efficient Transformer-based large-scale language representations using block structured pruning. The contributions of this work are as follows. 
\leftmargini=4mm
\begin{itemize}
\item To the best of our knowledge, we are the first to investigate \textit{hardware-friendly} weight pruning on pre-trained large-scale language models. Besides the significantly reduced weight storage and computation, the adopted block structure pruning has high flexibility in achieving a high compression rate. The two advantages are critical for efficient Transformer in NLP since the non-uniformed syntax and semantics information in language/text domain makes weight pruning more difficult than computer vision.

\item We incorporate the reweighted group Lasso for optimization into block structured pruning-based on pre-trained large-scale language models including BERT, RoBERTa, and DistilBERT. We relax the hard constraints in weight pruning by adding regularization terms in the objective function and use reweighted penalty parameters for different blocks. The dynamical regularization technique achieves higher compression rate with zero or minor accuracy degradation.

\item Our proposed method is orthogonal to existing compact pre-trained language models such as DistilBERT using knowledge distillation. We can further reduce the model size using our method with zero or minor accuracy.

\end{itemize}

We evaluate the proposed approach on several GLUE benchmark tasks~\cite{wang2018glue}. Experimental results show that we achieve high compression rates with zero or minor accuracy degradation.  With significant gain in weight storage reduction (up to 5$\times$) and computation efficiency, our approach can maintain comparable accuracy score to original large models including DistilBERT. The hardware-friendly transformer-based acceleration method is suitable to be deployed on resource-constrained edge devices.

\section{Related Work}

To address the memory limitation and high computational requirement of commonly seen deep learning platforms such as graphics processing unit (GPU), tensor processing unit (TPU) and field-programmable gate array (FPGA) on large-scale pre-trained language models, various of compact NLP models or model compression techniques have been investigated. ALBERT~\cite{lan2019albert} utilizes \textit{parameter sharing} technique across encoders to reduce weight parameters and uses the same layer structures as BERT. It achieves comparable results on different benchmarks to BERT. Despite the weight storage reduction, the computational overhead remains unchanged since ALBERT and BERT have the same network structure.

\begin{figure*}[ht]
\centering
\includegraphics[scale=0.25]{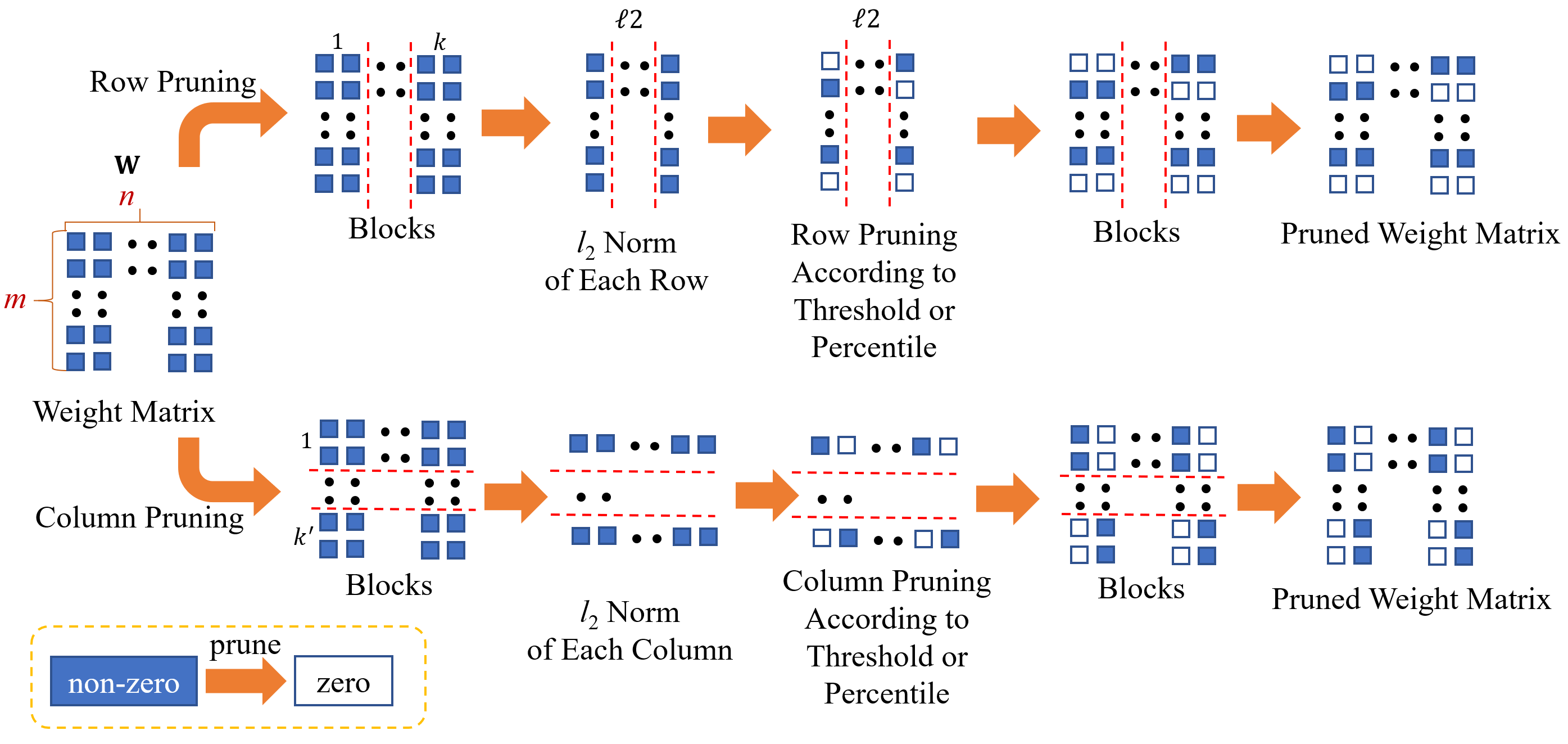}
\caption{Block structured pruning for weight matrix.}
\label{fig:BSPrune}
\end{figure*}

\textit{Knowledge distillation} is another type of model compression technique, which distills the knowledge from a large teacher model or an ensemble of models to a light-weighted student model \cite{hinton2015distilling}. 
The student model is trained to intimate the class probabilities produced by the large teacher model. 
For instance, DistilBERT~\cite{sanh2019distilbert} applies knowledge distillation to BERT, and achieves 1.67 $\times$ model size reduction and 1.63 $\times$ inference speedup, while retaining 97\% accuracy on the dev sets on the GLUE benchmark, compared to BERT. 
Patient knowledge distillation~\cite{sun2019patient} is used to learn from multiple intermediate layers of the teacher model for incremental knowledge extraction. 

Efficient deep learning methods can reduce the model size and accelerate the computation.  It is well known that, in practice, the weight representation in deep learning models is redundant.
After removing several redundant weights with appropriate model compression algorithms, the deep learning model can have minor accuracy degradation. Prior work focused on heuristic and iterative \textit{non-structured magnitude weight pruning} (a.k.a, irregular pruning)~\cite{han2015learning}. 
It causes overhead in both weight storage and computation in computer systems. On weight storage, it results in 
{\em irregular, sparse weight matrices} (as arbitrary weights can be pruned), and relies on indices to be stored in a compressed format such as Coordinate (COO) format. The introduced indices cause extra memory footprint, i.e., at least one index per non-zero value, further degrading the compression rate. 
On computation, it is difficult to be accelerated on current GPU architectures as reported in~\cite{han2016eie,wen2016learning,yu2017scalpel}. On the other hand, \textit{structured pruning} considers regularity in weight pruning focusing on generating regular but smaller and dense matrix with no index.  However,  it  suffers notable accuracy loss due to the poor solution quality, and therefore not suitable for pruning sensitive syntax and semantics information in Transformer.

\section{Block Structured Pruning}

\subsection{Problem Formulation}
{
We adopt a more fine-grained block structured pruning algorithm, where pruning is executed by excluding entire blocks of weights within weight matrices such as rows or columns, therefore significantly reducing the number of indices when storing on memory. On computation, it is compatible with  parallel computing platforms such as  GPUs or Field Programmable Gate Arrays (FPGAs) in implementing matrix multiplications. We formulate the weight pruning problem using reweighted group Lasso, to orchestrate the block structured pruning. 
Thus, the Transformer-based large-scale models can be more efficient on computer systems while satisfying the accuracy requirement. 
As shown in Figure~\ref{fig:BSPrune}, 
we divide the weight matrix into small \emph{blocks} and apply \emph{row pruning} and \emph{column pruning} on each block. For each row/column block, we compute the $l_2$ norm. {We prune the weights within the block according to our pre-set threshold or percentile.} The pseudocode is shown in Algorithm~\ref{alg:block_pruning}.

Consider an $N$-layer Transformer, we denote the weights and biases of the $n$-th layer as $\mathbf{W}_n$ and $\mathbf{b}_n$. The loss function is $f \big( \{{\mathbf{W}}_{n}\}_{n=1}^N, \{{\mathbf{b}}_{n} \}_{n=1}^N \big)$, which will be minimized during training. For the block structured pruning problem, our target objective is to reduce the number of columns and rows in the blocks of weight matrix while maintaining the prediction accuracy. 

\begin{equation}
\small
\label{original}
\begin{aligned}
 \text{minimize}&\quad  f \big( \{{{\bf W}_n}
\}_{n=1}^N, \{{{\bf b}_n}
\}_{n=1}^N \big)
\\ \text{subject to} 
&\quad\text{\# of non-zero block rows in ${\bf W}_n$ is less than $r_n$} \\
&\quad\text{\# of non-zero block columns in ${\bf W}_n$ is less than $c_n$}
\end{aligned}
\end{equation}

\begin{algorithm}[t]
\footnotesize
  \caption{Block structured pruning 
  }
  \label{alg:block_pruning}
\begin{algorithmic}
  \STATE {\bfseries Input:} weight matrix $\mathbf{W}$, matrix width $n$, matrix height $m$, row division $k$ (or column division $k'$), threshold $t_b$
  \STATE {\bfseries Output:} pruned weight matrix $\mathbf{W}_p$
  \STATE Set $\mathbf{W}_p$ = $\mathbf{W}$
  \STATE Divide $\mathbf{W}_p$ into $k$ matrices: $\mathbf{W}_{1}$,$\mathbf{W}_{2}$,...,$\mathbf{W}_{k}$
  \STATE Set $l_2\_norms$ = $zeros(k,m)$
  \FOR{$i=1$ to $k$ }
    \FOR{$j=1$ to $m$ }
        \STATE $l_2\_norms(i,j)$ equals the $l_2$ norm of the $j$ th row of $\mathbf{W}_{i}$ 
        \IF {$l_2\_norms(i,j) \le t_b$} 
            \STATE $\mathbf{W}_{i}$($j$,:) = 0
        \ENDIF 
    \ENDFOR        
  \ENDFOR
  \STATE $\mathbf{W}_p$ = concatenate($\mathbf{W}_{1}$,$\mathbf{W}_{2}$,...,$\mathbf{W}_{k}$)
\end{algorithmic}
\end{algorithm}



 

\noindent where $r_n$ and $c_n$ are the desired non-zero block rows and columns, respectively.
Due to regularity in pruning, only the non-zero rows/columns at the block level need to be indexed, as opposed to each non-zero element in irregular pruning. The storage overhead is minor compared to non-structured irregular pruning~\cite{han2016eie}.
Because structured pruning is applied independently within each block, the scheme has higher flexibility, thereby higher accuracy, compared to the straightforward application on the whole weight matrix~\cite{wen2016learning}. 
}

\subsection{Reweighted Group Lasso Optimization}

In problem (\ref{original}), we use hard constraints to formulate the block row/column pruning problem. However, it is more difficult to satisfy the hard constraints on NLP than on computer vision. 
There are two reasons: i) Information compressed in image features can be partially retrieved from neighboring pixels since spatially they share similar and uniform characteristics, whereas syntax and semantics information in deep Transformer in language/text domain are not uniformly characterized; ii) Intuitively, the high-level semantic, syntax, and language understanding capability might be broken when we
prune zero or near-zero weights in the latent space. Therefore, a high compression rate for large-scale language models is difficult to achieve on downstream NLP tasks.

To address this issue, we relax the hard constraints by adding regularization terms in the objective function. Prior work SSL~\cite{wen2016learning} uses group Lasso as the relaxation of the hard constraints. Inspired by \cite{candes2008enhancing}, we use reweighted penalty parameters for different blocks to achieve a high compression rate under same accuracy requirement than using a fixed penalty parameter to all the blocks in group Lasso method.

When we use group Lasso for block row pruning, the regularization term is 
\[
\sum_{n=1}^{N} \sum_{i=1}^{p_n} \sum_{\alpha=1}^{q_n} \sqrt{\sum_{j=(\alpha-1)h_n+1}^{\alpha h_n}(\mathbf{W}_n)_{ij}^2}
\]
where $h_n$ is the block row size in the $n$-th layer, $p_n$ is the number of rows in $\mathbf{W}_n$, $q_n$ is the number of blocks in a row of $\mathbf{W}_n$.
And the block row pruning problem is
\begin{equation}
\begin{aligned}
 & \underset{ \{{{\mathbf{W}}}_{n}\},\{{\mathbf{b}}_{n} \}}{\text{min}}
\ \ \  f \big( \{{{\mathbf{W}}}_{n} \}_{n=1}^N, \{{{\mathbf{b}}}_{n} \}_{n=1}^N \big)\\ & +  \lambda\sum_{n=1}^{N} \sum_{i=1}^{p_n} \sum_{\alpha=1}^{q_n} \gamma_{i,\alpha}\sqrt{\sum_{j=(\alpha-1)h_n+1}^{\alpha h_n}(\mathbf{W}_n)_{ij}^2}, \\
\end{aligned}
\end{equation}
where $\lambda$ is the penalty parameter. $\gamma_{i,\alpha}$ is the penalty weight corresponding to the $\alpha$-th block in the $i$-th row, and it is updated by 
$\gamma_{i,\alpha} = {1}/({\sqrt{\sum_{j=(\alpha-1)h_n+1}^{\alpha h_n}({\bf{\mathbf{W}}}_n)_{ij}^2}+\epsilon})$, where $\epsilon$ is a small value preventing division by zero.
Similarly, when we prune columns in a block, the problem becomes 
\begin{equation}
\begin{aligned}
& \underset{ \{{\mathbf{W}}_{n}\},\{{\mathbf{b}}_{n} \}}
{\text{min}}
\ \ \ f \big( \{{\mathbf{W}}_{n} \}_{n=1}^N, \{{\textbf{{b}}}_{n} \}_{n=1}^N \big) \\ 
& +\lambda\sum_{n=1}^{N} \sum_{j=1}^{r_n} \sum_{\beta=1}^{s_n} \gamma_{j,\beta}
\sqrt{\sum_{i=(\beta-1)d_n+1}^{\beta d_n}(\textbf{W}_n)_{ij}^2}, \\
\end{aligned}
\label{eq:mixed_loss}
\end{equation}
where $d_n$ is the block column size in the $n$-th layer, $r_n$ is the number of columns in $\textbf{W}_n$. $s_n$ is the number of blocks in a column of $\textbf{W}_n$. $\gamma_{j,\beta}$ is the penalty weight corresponding to the $\beta$-th block in the $j$-th column, and it is updated by 
$\gamma_{j,\beta} = {1}/({\sqrt{\sum_{i=(\beta-1)d_n+1}^{\beta d_n}(\mathbf{W}_n)_{ij}^2}+\epsilon}).$

\begin{algorithm}[t]
\footnotesize
  \caption{Reweighted group Lasso on Transformer pruning}
  \label{alg:example2}
\begin{algorithmic}
  \STATE {\bfseries Input:} pre-trained model, model weight matrix $\mathbf{W}$, matrix width $n$, matrix height $m$
  \STATE Set milestones = {$m_{1}$, $m_{2}$, ..., $m_{s}$}
  \STATE Set $T_{1}$ as the number of iterations of reweighted training method
  \STATE Set $T_{2}$ as the number of iterations of retraining method
  \STATE Calculate $\gamma$
  \FOR{$s=1$ to $T_{1}$ }
        \IF {$s$ in milestones} 
            \STATE Update $\gamma$
        \ENDIF 
        \STATE Calculate $l_{1loss}$ and prediction loss $f(\textbf{W},\textbf{b})$
        \STATE $mixed_{loss} = l_{1loss} + f(\textbf{W},\textbf{b})$
        \STATE Update model weight $\mathbf{W}$ to minimize $mixed_{loss}$ using Adam
  \ENDFOR
  \STATE Prune the weight matrix $\mathbf{W}$ using block structured pruning
  \STATE $Mask = zeros(m,n)$
  \FOR{$i=1$ to $m$ }
        \FOR{$j=1$ to $n$ }
     
            \IF {$\textbf{W}_{i,j} == 0$} 
                \STATE Set $Mask_{i,j}$ = 0
            \ELSE
                \STATE Set $Mask_{i,j}$ = 1
            \ENDIF
        
        \ENDFOR
  \ENDFOR    
  \FOR{$s=1$ to $T_{2}$ }
        \STATE Calculate the prediction loss $f(\mathbf{W},\mathbf{b})$
        \STATE Update model weight $\mathbf{W}$ to minimize $f(\textbf{W},\textbf{b})$ using Adam
        \STATE $\textbf{W} = \textbf{W}*Mask$
  \ENDFOR  
\end{algorithmic}
\end{algorithm}


We start with a pre-trained model and initialize the collection of penalty weights ($\gamma_{i,\alpha}$ or $\gamma_{j,\beta}$) using the parameters
in the pre-trained model.
 We remove the rows or blocks in a block if their group $l_2$ norm is smaller than a threshold after reweighted training. We refine the Transformer models using the non-zero weights.
$\lambda$ is used for adjusting regularization strength. When $\lambda$ is too small, the reweighted training is close to the original training. When $\lambda$ is too large, it gives too much penalty on the weights and accuracy cannot be maintained. Specifically, we start reweighted training with $\lambda=0$ to reproduce the original results and increase $\lambda$ to derive sparsity of the weight matrices.  We stop increasing $\lambda$ when the reweighted training accuracy drops slightly and the accuracy will be improved after retraining. Overall, using the same training trails, our method can achieve higher pruning rate than prior works using structured pruning~\cite{wen2016learning}, as shown in Algorithm~\ref{alg:example2}.

\section{Evaluation}

\subsection{Datasets}
We conduct experiments on GLUE benchmark \cite{wang2018glue}, a comprehensive collection of nine natural language understanding tasks covering three NLP task categories with different degrees of difficulty and dataset scales: single-sentence tasks, paraphrase similarity matching tasks, and inference tasks. All datasets are public available.
More specifically, for single-sentence task, we consider the Corpus of Linguistic Acceptability (CoLA) \cite{warstadt2018neural}, which contains 10,657 sentences of English acceptability judgments from books and articles on linguistic theory, and the Stanford Sentiment Treebank (SST-2) \cite{socher-etal-2013-recursive}, which is comprised of 215,154 phrases in the parse trees of 11,855 sentences from movie reviews with annotated emotions.

For paraphrase similarity matching tasks, we consider the Microsoft Research Paraphrase Corpus (MRPC) \cite{dolan2005automatically}, which contains 5,800 sentence-pairs corpora from online news sources and are manually annotated where the sentences in the sentence-pairs are semantically equivalent; the Semantic Textual Similarity Benchmark (STS-B) \cite{cer-etal-2017-semeval}, a collection of 8,628 sentence pairs extracted from the news title, video title, image title, and natural language inference data; and the Quora Question Pairs (QQP) \footnote{https://www.quora.com/q/quoradata/First-Quora-Dataset-Release-Question-Pairs}, a collection of 400,000 lines of potential question duplicate pairs from the Quora website. 

For inference tasks, we consider the Multi-Genre Natural Language Inference Corpus (MNLI) \cite{N18-1101}, a set of 433k premise hypothesis pairs to predict whether the premise statement contains assumptions for the hypothesis statement; Question-answering NLI (QNLI) \cite{wang2018glue}, a set of over 100,000+ question-answer pairs from SQuAD \cite{DBLP:journals/corr/RajpurkarZLL16}; 
The Recognizing Textual Entailment datasets (RTE) \cite{wang2018glue}, which come from the PASCAL recognizing  Textual Entailment Challenge; and Winograd NLI (WNLI) \cite{wnli}, a reading comprehension task that comes from the Winograd Schema Challenge. 

In all GLUE benchmarks, we report the metrics following the conventions in \cite{wang2018glue}, i.e., accuracy scores are reported for SST-2, QNLI, RTE, and WNLI; Matthews Correlation Coefficient (MCC) is reported for CoLA; F1 scores are reported for QQP and MRPC; Spearman correlations are reported for STS-B.

\subsection{Experimental Setup}

\textbf{Baseline Models.}
Our baseline models are unpruned BERT$_{\mathrm{BASE}}$~\cite{devlin2018bert}, RoBERTa$_{\mathrm{BASE}}$~\cite{liu2019roberta}, and DistilBERT~\cite{sanh2019distilbert}. As shown in Table~\ref{table:main}, for each transformer model, we list the reported accuracy/metrics from the original papers in the first row. We report our reproduced results using the same network architectures in the second row.

\begin{table*}[!h]
	\centering
	\caption{{Comparison of test accuracy using different transformer models among the nine GLUE benchmark tasks.}}\label{table:main}
	\resizebox{1\textwidth}{!}{
		\begin{tabular}{l| l l l l l l l l l}
			\hline
			Models & MNLI & QQP & QNLI & SST-2 & CoLA & STS-B & MRPC & RTE & WNLI\\
\hline
\textbf{BERT$_{\mathrm{BASE}}$ \cite{devlin2018bert}} & 84.6 & 91.2 & 90.5 & 93.5 & 52.1 & 85.8 & 88.9 & 66.4 & - \\
\textbf{BERT$_{\mathrm{BASE}}$ (ours)} & 83.9 & 91.4 & 91.1 & 92.7 & 53.4 & 85.8 & 89.8 & 66.4 & 56.3 \\
\textbf{BERT$_{\mathrm{BASE}}$ prune (ours)} & 82.9 & 90.7 & 88.2 & 89.3 & 52.6 & 84.6 & 88.3 & 63.9 & 56.3\\
\textbf{Compression rate} & 1.428$\times$ & 1.428$\times$ & 1.428$\times$ & 1.428$\times$ & 1.428$\times$ & 1.428$\times$ & 1.428$\times$ & 1.428$\times$ & 2.0$\times$\\
\hline

\textbf{RoBERTa$_{\mathrm{BASE}}$ \cite{liu2019roberta}} & 87.6 & 91.9 & 92.8 & 94.8 & 63.6 & 91.2 & 90.2 & 78.7 & -	 \\
\textbf{RoBERTa$_{\mathrm{BASE}}$ (ours)} &87.8 & 91.6 & 93.0 & 94.7 & 60.1 & 90.2 & 91.1 & 77.3 & 56.3 \\
\textbf{RoBERTa prune (ours)} & 86.3 & 87.0 & 90.0 & 89.2 & 55.3 & 88.8 & 90.2& 74.0 & 56.3 \\
\textbf{Compression rate} & 1.428$\times$ & 1.428$\times$ & 1.428$\times$ & 1.428$\times$ & 1.246$\times$ & 1.428$\times$ & 1.428$\times$ & 1.428$\times$ & 2.0$\times$\\
\hline
\textbf{DistilBERT \cite{sanh2019distilbert}} & 82.2 & 88.5 & 89.2 & 91.3 & 51.3 & 86.9 & 87.5 & 59.9 & 56.3\\
\textbf{DistilBERT (ours)} & 81.9 & 90.2 & 89.5 & 90.9 & 50.7 & 86.5 & 89.8 & 59.2 & 56.3\\
\textbf{DistilBERT prune (ours)} & 78.5 & 87.4 & 85.3 & 85.3 & 53.4 & 83.7 & 89.1 & 59.2 & 56.3 \\
\textbf{Compression rate} & 2.0$\times$ & 1.667$\times$ & 1.667$\times$ & 2.0$\times$ & 1.197$\times$ & 1.667$\times$ & 1.207$\times$ & 2.0$\times$ & 2.0$\times$ \\
\hline
		\end{tabular}
	}
\end{table*}

\noindent\textbf{Evaluation Metrics.}
To evaluate our proposed framework on NLP model compression problems, we apply our method on different transformer-based models including BERT$_{\mathrm{BASE}}$, RoBERTa$_{\mathrm{BASE}}$, and DistilBERT. Reweighted $l_1$ training is carried out to add $l_1$ regularization, block structured pruning to obtain a sparse model, and retraining to improve the final accuracy. 

{We access the GPU-AI (Bridges GPU Artificial Intelligence) nodes on the Extreme Science and Engineering Discovery Environment (XSEDE)~\cite{xsede}. We use two node types: Volta 16 - nine HPE Apollo 6500 servers, each with 8 NVIDIA Tesla V100 GPUs with 16 GB of GPU memory each, connected by NVLink 2.0; Volta 32 - NVIDIA DGX-2 enterprise research AI system tightly coupling 16 NVIDIA Tesla V100 (Volta) GPUs with 32 GB of GPU memory each, connected by NVLink and NVSwitch.
We also use an  $8\times$ NVIDIA Quadro RTX 6000 GPU server with 24 GB of GPU memory each for training. 
We conduct the experiments using HuggingFace Transformer toolkit for the state-of-the-art NLP \cite{wolf2019huggingface} and the \textit{DeepLearningExamples} repository from NVIDIA ~\cite{2020_deeplearningexamples}}.
Our experiments are performed on Python 3.6.10, GCC 7.3.0, PyTorch 1.4.0, and CUDA 10.1.

We show the prediction accuracy with respect to different compression rates and we evaluate our method on the GLUE benchmark \cite{wang2018glue} in Table~\ref{table:main}. 
For BERT, we use the official BERT$_{\mathrm{BASE}}$, uncased model as our pre-trained model. There are 12 layers ($L$ =12; hidden size $H$ = 768; self-attention heads $A$ = 12), with total number of parameters 110 Million. We use the same fine-tuning hyperparameters as the paper \cite{devlin2018bert}. 
For RoBERTa \cite{liu2019roberta}, we use the official RoBERTa$_{\mathrm{BASE}}$ model as our pre-trained model. It has the same structure as the BERT$_{\mathrm{BASE}}$ model, with 12 layers ($L$=12; hidden size $H$= 768; self-attention heads $A$= 12), and a total number of 125 Million parameters.
For DistilBERT \cite{sanh2019distilbert}, a distilled model from the BERT$_{\mathrm{BASE}}$, uncased checkpoint, is used as the pre-trained model. The parameters are $L$ = 6; $H$ = 768; $A$ = 12; total parameters = 66 M. The block size used for pruning has different types, e.g., 3$\times$3, 3$\times$12, and 12$\times$3.

\subsection{Implementation Details}
We first fine-tune the pre-trained models for classification. BERT, RoBERTa, and DistilBERT share the same steps. We add a single linear layer on top of each original model. We train the model for the nine downstream GLUE tasks with their corresponding datasets. As we feed the data, the entire pre-trained model and the additional untrained classification layer is trained on our specific task. The original layers already have great English words representation, and we only need to train the top layer, with a bit of tweaking in the lower levels to accommodate our task.

For fine-tuning, we run 4 epochs with initial learning rate of $2e^{-5}$,  batch size of 32 and warm up proportion of 0.1. For block structured pruning, we adjust the reweighted penalty parameter, compression rate and training steps for each task. We use the same parameters as fine-tuning (epochs, learning rate, batch size), then we adjust some parameters for each task, depending on the prediction performance.
For BERT$_{\mathrm{BASE}}$, we set penalty factor $1e^{-3}$ for WNLI and MRPC; penalty factor $1e^{-4}$ for CoLA, QQP, MNLI, SST-2, and RTE; penalty factor $1e^{-5}$ for QNLI. The learning rate is $3e^{-5}$ and batch size is 32 on nine tasks.
For RoBERTa$_{\mathrm{BASE}}$, we set penalty factor $1e^{-3}$ for WNLI; penalty factor $1e^{-4}$ for MRPC, QQP, SST-2, and RTE; penalty factor $1e^{-5}$ for QNLI, CoLA, and MNLI. 
The learning rate  and  batch size are the same as BERT$_{\mathrm{BASE}}$.
For DistilBERT model, the hyperparamters for reweighted training and retraining are learning rate = $3e^{-5}$ and batch size = 128 on nine datasets. We adjust other parameters, including penalty factors, number of blocks, and compression ratios to achieve the satisfied performance on each task.

\begin{figure*}[h]
\centering
\includegraphics[width=1.8\columnwidth]{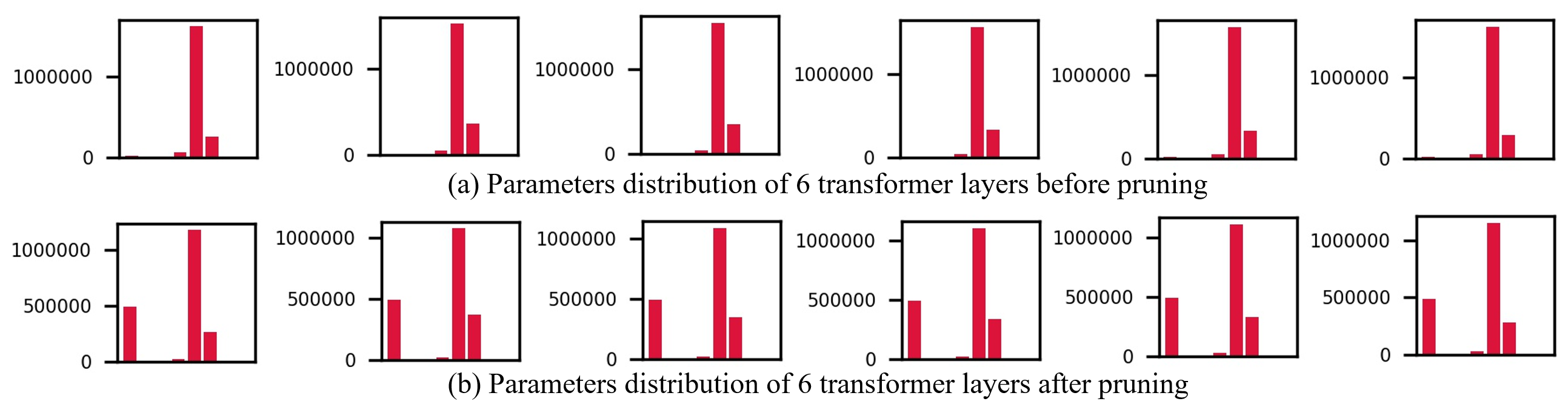}
\caption{Parameters distribution of DistilBERT model on CoLA dataset: (a) before pruning, (b) after pruning.}
\label{fig:Param_Dist_CoLA}
\end{figure*}

We consider three objectives: weight distribution, loss, and accuracy. Weight distribution shows the distribution of weights in each layer after training and retraining. We visualize the weight parameters in Figure~\ref{fig:Param_Dist_CoLA}.
With different pruning hyper-parameters including penalty factors, learning rate, block numbers, and compression rate, the weights are distributed differently. We look at two losses: reweighted loss and mixed loss (the object function in Equation \eqref{eq:mixed_loss}). 
For all our tasks, BERT$_{\mathrm{BASE}}$, RoBERTa$_{\mathrm{BASE}}$, and DistilBERT are converged in less than 4 epochs. During training, we evaluate the performance between each given steps. 

\subsection{Experimental Results}

\begin{figure}[t]
\includegraphics[width=0.95\columnwidth]{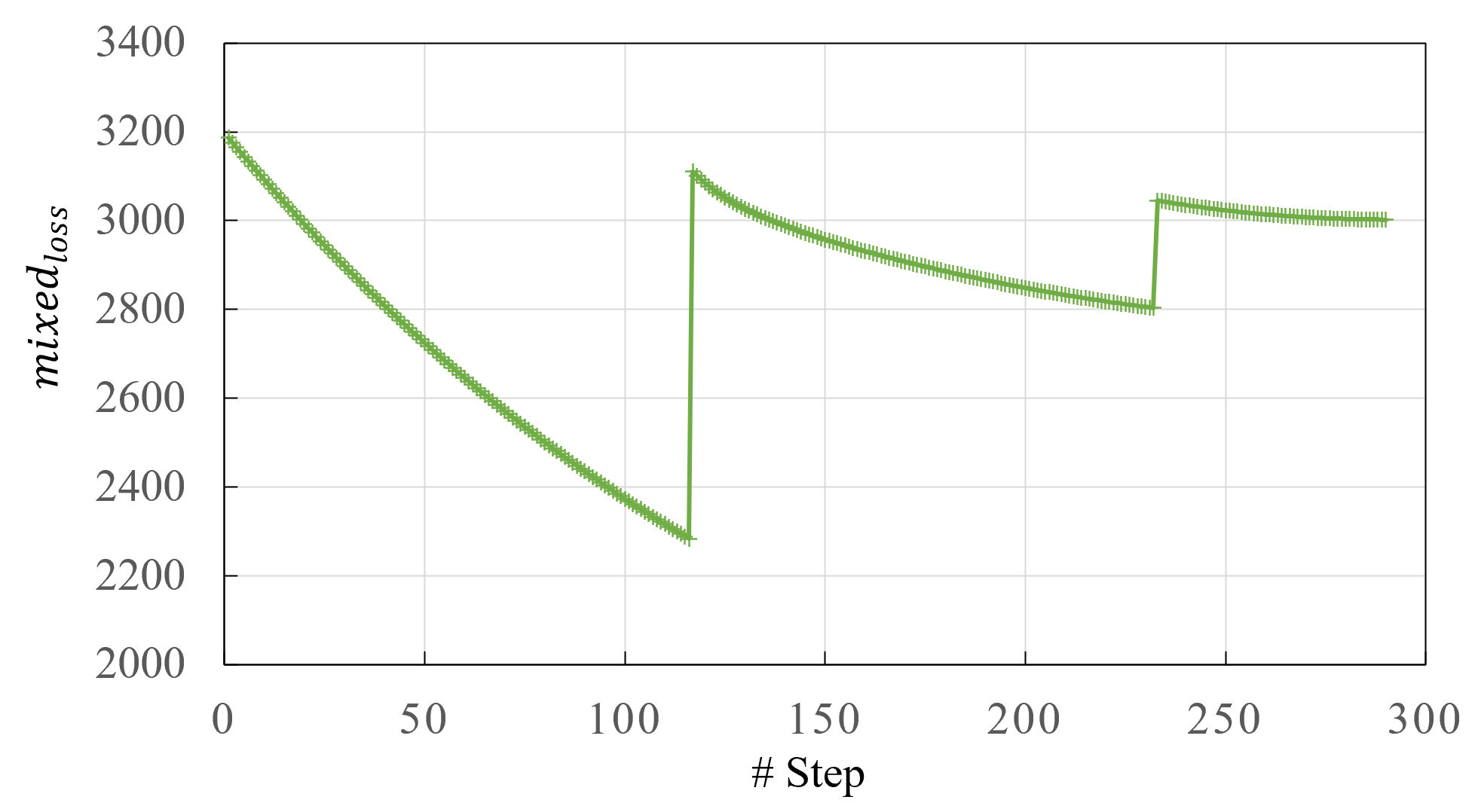}
\caption{Mixed loss of reweighted training on MRPC dataset with DistilBERT model.}
\label{fig:MRPC_mixed_loss}
\end{figure}

\begin{figure}[b]
\includegraphics[width=1\columnwidth]{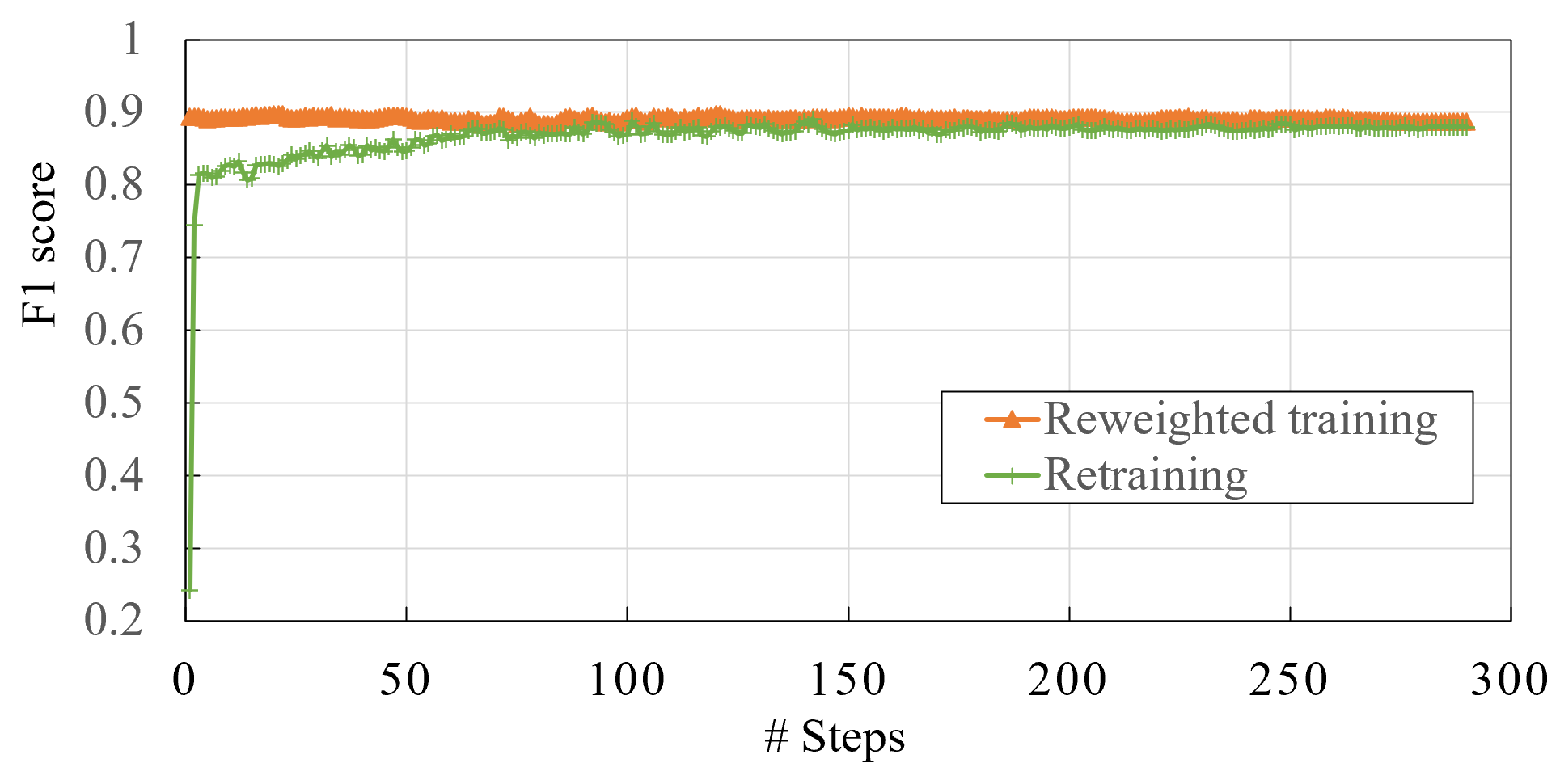}
\caption{F1 score of reweighted training and retraining with DistilBERT model on MRPC dataset.}
\label{fig:Distilbert_MRPC_acc}
\end{figure}

We compare the performance (accuracy score) of our pruned models with the baselines. The results are shown in Table~\ref{table:main}. 
For BERT$_{\mathrm{BASE}}$, we set a compression rate of 1.428$\times$ (i.e., 30\% sparsity) or above. The average accuracy degradation is within 2\% on all tasks.
On WNLI task, there is no accuracy loss. On MNLI, QQP, CoLA, STS-B, and MRPC tasks, the accuracy loss is within 1.5\%. On SST-2, QNLI, and RTE tasks, the accuracy loss is also small (within 2.9\%), compared to two baseline models.
For RoBERTa, the average accuracy degradation is within 3\% on all tasks. There is no accuracy loss on WNLI. The accuracy loss is within 1\% on MRPC, within 2\% on MNLI and STS-B tasks, within 4\% on QNLI and RTE tasks, around 5\% on QQP, SST-2 and CoLA tasks.
 For DistilBERT, the average accuracy degradation is within 5\% on all tasks. The accuracy losses are within 1\% on MRPC task, 3\% on MNLI, QQP, QNLI, and STS-B tasks, and 5\% on SST-2 task. On CoLA and WNLI datasets, the pruned models perform even better than the unpruned models and increase the final accuracy by 3\% (1.197$\times$ compression rate) and 4\% (2.0$\times$ compression rate), respectively.
  Figure~\ref{fig:MRPC_mixed_loss} and Figure~\ref{fig:Distilbert_MRPC_acc} show
the reweighted training and retraining results on MRPC dataset, respectively. We choose 256 as the number of blocks. For reweighted training, the mixed loss drops during training within every 116 steps (4 epochs) and increases significantly since we update the penalty matrix $\gamma$. For retraining, the pruned model achieves higher F1 score than the unpruned one.

\begin{table}[t]
	\centering
	\caption{Pruning results of BERT$_{\mathrm{BASE}}$ with different compression rates.}
	\label{table:sparsity_results}
	\resizebox{1.0\columnwidth}{!}{
		\begin{tabular}{c| c c c c c}
			\hline
            Compression rate & QQP &  MNLI  & WNLI &QNLI & SST-2 \\
            \hline
            1$\times$ & 91.4 &  83.9 & 56.3 &91.1 & 92.7 \\
            1.428$\times$ & 90.7 &  82.9 &  56.3 & 88.2& 89.3\\
            2.0$\times$ & 90.0 &  81.2 &  56.3 &85.5 &87.0\\
            5.0$\times$ & 86.9  & 76.6 & 56.3 &79.5 & 82.3\\
            \hline
        \end{tabular}
        }
\end{table}

We evaluate the accuracy changes when compression rates are different on BERT$_{\mathrm{BASE}}$ and report the accuracy scores for different tasks. Results indicate that the sensitivities of tasks vary significantly under different levels of compression rates. As shown in Table~\ref{table:sparsity_results}, different tasks show different accuracy degradation when using the same compression rate. As we increase the compression rate, the accuracy degradation increased.
For specific task (e.g., WNLI), we can achieve up to 5$\times$ compression rate from baseline model with zero accuracy loss. Results on tasks such as WNLI and QQP show minor accuracy degradation while results on SST-2, MNLI, QNLI, show higher accuracy degradation when compression rate is 5.0$\times$. The different accuracy results are related to different dataset sizes, degrees of difficult, and evaluation metrics. 

\begin{table}[t]
\centering
	\caption{Comparison of test accuracy between our BSP method and irregular sparse format on GLUE benchmarks.}\label{table:irregular_sparse}
	\resizebox{1.0\columnwidth}{!}{
\begin{tabular}{l|l l l l l l l}
\hline
Network Model    & MNLI & QQP & QNLI & SST2 & SSTB & RTE & WNLI \\ \hline
\begin{tabular}[c]{@{}l@{}}BERT$_{\mathrm{BASE}}$ prune\end{tabular}& 82.9 & 90.7 & 88.2 & 89.3 & 84.6 & 63.9 & 56.3\\ 
\begin{tabular}[c]{@{}l@{}}Prune ratio\end{tabular}& 1.428$\times$ & 1.428$\times$ & 1.428$\times$ & 1.428$\times$ & 1.428$\times$ & 1.428$\times$ & 2.0$\times$\\ \hline
\begin{tabular}[c]{@{}l@{}}BERT$_{\mathrm{BASE}}$ irregular\end{tabular}& 83.7 & 86.5 & 87.8 & 90.8 &  86.7 & 63.5 & 56.3 \\ 
\begin{tabular}[c]{@{}l@{}}Prune ratio\end{tabular}&  2.0$\times$ & 2.0$\times$ & 1.667$\times$ & 2.0$\times$ & 2.5$\times$ & 2.373$\times$ & 2.0$\times$ \\ \hline
\begin{tabular}[c]{@{}l@{}}DistilBERT prune\end{tabular}& 78.5 & 87.4 & 85.3 & 85.3 &  83.7 & 59.2 & 56.3 \\ 
\begin{tabular}[c]{@{}l@{}}Prune ratio\end{tabular}& 2.0$\times$ & 1.667$\times$ & 1.667$\times$ & 2.0$\times$ & 1.667$\times$ & 2.0$\times$ & 2.0$\times$ \\ \hline
\begin{tabular}[c]{@{}l@{}}DistilBERT irregular\end{tabular}& 80.3 & 88.7 & 87.2 & 86.7 &  84.7 & 59.9 & 56.3 \\ 
\begin{tabular}[c]{@{}l@{}}Prune ratio\end{tabular}& 2.381$\times$ & 2.174$\times$ & 2.326$\times$ & 2.222$\times$ & 2.222$\times$ & 2.083$\times$ & 2.0$\times$ \\ \hline


\end{tabular}
}
\end{table}

We compare our BSP method with irregular sparse format and the block sparse format~\cite{narang2017blocksparse,Gray2017GPUKF} (pruning all weights on selected blocks). Table~\ref{table:irregular_sparse} shows that under same accuracy, our method achieves a slightly lower pruning ratio compared to irregular sparse format. This is because irregular pruning has a larger flexibility in pruning. However, irregular pruning is less effective when applying to hardware. Irregular sparse format introduces significant memory storage overhead when using Coordinate Format (COO) storage format,
therefore is not hardware-friendly.  Our method, block structured format (pruning a portion of rows/columns on each block) strikes a better balance between accuracy and memory storage than irregular sparse format or block sparse format~\cite{narang2017blocksparse,Gray2017GPUKF}. For irregular sparse format, when storing or transmitting an irregular sparse matrix using the COO format, we store the subsequent nonzeros and related coordinates in memory. Three vectors are needed: row, col, data, where data[i] is value at (row[i], col[i]) position.
More specifically, given $50\%$ sparsity for a $8\times8$ matrix with block size of 4$\times$4, the storage of COO format is $1.5\times8\times8=96$; the storage of block structured sparsity (row only) is $8*8\times0.5+16$
=48. 
Table~\ref{table:block_sparse} lists the accuracy of our method and block sparse format using DistilBERT.
Our method achieves $3.04\%$ higher accuracy in average compared with block sparse format.

\begin{table}[t]
\centering
	\caption{Comparison of test accuracy between our BSP method and block sparse method~\cite{narang2017blocksparse} on GLUE benchmarks.}\label{table:block_sparse}
	\resizebox{1.0\columnwidth}{!}{
\begin{tabular}{l|l l l l l l l}
\hline
Network Model    & MNLI & QQP & QNLI & SST2 & SSTB & RTE & WNLI \\ \hline
\begin{tabular}[c]{@{}l@{}}DistilBERT \end{tabular}& 81.9 & 90.2 & 89.5 & 90.9 &  86.5 & 59.2 & 56.3\\ \hline
\begin{tabular}[c]{@{}l@{}}DistilBERT-prune\end{tabular}& 78.5 & 87.4 & 85.3 & 85.3 &  83.7 & 59.2 & 56.3 \\ \hline
\begin{tabular}[c]{@{}l@{}}Block Sparse\end{tabular}&78.3 & 87.2 &85.2 & 83.9 &82.2 & 58.8 & 49.3 \\ 
\hline
\begin{tabular}[c]{@{}l@{}}Accuracy Loss\end{tabular} &0.2  & 0.2 & 0.1 & 1.4& 1.5 & 0.4 &  13 \\ 
\hline 

\end{tabular}
}
\end{table}

As the proposed pruning is hardware-friendly, the pruned weights can be efficiently stored in hardware memory with minor overhead  (compared to other pruning methods like irregular pruning). 
We use a compiler-assisted acceleration framework other than sparse linear algebra libraries, which allows the model to speed up with a sparsity of 30\%. We also apply matrix reorder and compact model storage to achieve speed up on edge devices~\cite{ma2020blkrew}.
Hence, it is suitable to deploy the final compressed model on resource-constrained edge devices such as embedded systems and mobile devices. 

\section{Ablation Studies}

In this section, we perform ablation experiments over several parameters when pruning BERT and DistilBERT to better understand their relative importance and the procedure. We change the selection of following parameters: the numbers of blocks for reweighted training and block structured pruning, retraining epochs, and penalty factors. We also evaluate the knowledge distillation friendly.

\subsection{Number of Blocks}
After selecting penalty factor $3e^{-4}$ and compression rate 2.0$\times$ for each layer (except embedding layers), we choose different numbers of blocks to test. As shown in Table~\ref{table:ablation_results_CoLA}, the final accuracy is significantly improved for both BERT$_{\mathrm{BASE}}$ and DitilBERT when we increase the number of blocks. It verifies that with more number of blocks (smaller block size), our weight pruning algorithm has higher flexibility in exploring model sparsity. 

\begin{table}[!ht]
	\centering
	\vskip 0.1cm
	\caption{Number of blocks for reweighted training and retraining on CoLA dataset.}
	\label{table:ablation_results_CoLA}
	\resizebox{1.0\columnwidth}{!}{
		\begin{tabular}{l| l l l l}
			\hline
            Number of blocks & 8 & 128 & 256 & 768 \\
            \hline
            BERT$_{\mathrm{BASE}}$ retraining MCC & 14.5 & 48.0 & 52.6 & 51.5 \\
            DistilBERT retraining MCC & 32.2 & 43.8 & 47.2 & 53.4 \\

            \hline
        \end{tabular}
	}
\end{table}

\subsection{Number of Retraining Epochs}
By default, all GLUE tests are carried out by running four epochs for pre-training. For reweighted training and retraining, more epochs usually lead to better final accuracy. In this test, we try different reweighted training and retraining epochs. During reweighted training, the mixed loss will drop significantly within every 4 epochs, while the evaluation loss keeps relatively stable. 
We summarize the results in Table~\ref{table:ablation_results_STS-B}. The final accuracy of retraining is improved when we increase the training epochs.

\begin{table}[!ht]
	\centering
	\vskip 0.1cm
	\caption{Retraining epochs on STS-B dataset.}
	\label{table:ablation_results_STS-B}
	\resizebox{1.0\columnwidth}{!}{
		\begin{tabular}{l| l l l}
			\hline
            Number of epochs & 4 & 8 & 16 \\
            \hline
            BERT$_{\mathrm{BASE}}$ retraining Spearman  & 84.2 & 84.4 & 84.6  \\
            DistilBERT retraining Spearman  & 74.6 & 79.1 & 80.9  \\

            \hline
        \end{tabular}
	}
\end{table}

\subsection{Penalty Factors}
The reweighted training procedure is utilized to penalize the $l_{2}$ norm of the blocks and thus to reduce the magnitude of the weights. Therefore, larger penalty factors help to achieve better retraining accuracy since more smaller weight values of the weight matrices are pruned. However, if the penalty factors are too large, the reweighted training algorithm is not able to compress the model well, which leads to significant accuracy degradation. The results are summarized in Table~\ref{table:ablation_results_MNLI}. The retraining accuracy is improved when we increase the penalty factor from $3e^{-5}$ to $1e^{-4}$ and declines from $3e^{-4}$ to $1e^{-3}$.

\begin{table}[!ht]
	\centering
	\vskip 0.1cm
	\caption{Penalty selections on MNLI dataset.}
	\label{table:ablation_results_MNLI}
	\resizebox{1.0\columnwidth}{!}{
		\begin{tabular}{l| l l l l}
			\hline
            Penalty factor for each layer & $3e^{-5}$ & $1e^{-4}$ & $3e^{-4}$& $1e^{-3}$\\
            \hline
            BERT$_{\mathrm{BASE}}$ retraining accuracy  & 80.7 & 82.5 & 82.9 & 78.9   \\
            DistilBERT retraining accuracy  & 65.8 & 68.8 & 73.6 & 70.0  \\

            \hline
        \end{tabular}
	}
\end{table}

\subsection{Variance of results on multiple runs}
During our training, the random seeds are set to 42 as default. We further conduct experiments choosing different seeds and list the results in Table~\ref{table:multiple_runs}. We observe our reported accuracy is aligned with the results with different seeds.

\begin{table}[!ht]
	\centering
	\vskip 0.1cm
	\caption{Variance of results on multiple runs.}
	\label{table:multiple_runs}
	\resizebox{1.0\columnwidth}{!}{
		\begin{tabular}{l| l l l l}
			\hline
            Seed & SST-2  & CoLA & STS-B & MRPC\\
            \hline
            42(default)  & 85.3 & 53.4 & 83.7 & 89.1   \\
            1  & 83.14 & 53.75 & 83.19 & 89.3  \\
            1000  & 82.8 & 54.08 & 83.32  & 89.3  \\
            5000  & 82.91 & 54.22 & 83.03 & 89.0  \\

            \hline
        \end{tabular}
	}
\end{table}

\subsection{Knowledge Distillation Friendly}
To evaluate the effectiveness of our pruning method on distilled models, we focus on the BERT and DistilBERT results in Table \ref{table:main}, where DistilBERT is a highly distilled version of BERT. 
The average compression rate of BERT and DistilBERT are 1.49$\times$ and 1.79$\times$, respectively. 
Please note that model size of BERT is 1.67$\times$ of DistilBERT, and therefore is 2.99$\times$ of the final compressed DistilBERT model size. 
This show that the proposed block structured pruning is orthogonal to {knowledge distillation}.
With this \textit{knowledge distillation friendly} property, we can first apply the standard knowledge distillation step to reduce the original large model and then apply the proposed pruning method to further reduce the size of the student model.

\section{Conclusion}
In this work, we propose an hardware-friendly block structured pruning pruning framework for transformer-based large-scale language representation. We incorporate the reweighted group Lasso into for optimization and relax the hard constraints in block structured pruning.
We significantly reduce weight storage and computational requirement. Experimental results on different models (BERT, RoBERTa, and DistilBERT) on the GLUE benchmark tasks show that we achieve significant compression rates with zero or minor accuracy degradation on certain benchmarks. Our proposed method is orthogonal to existing compact pre-trained language models such as DistilBERT using knowledge distillation. It is suitable to deploy the final compressed model on resource-constrained edge devices.

\section*{Acknowledgement}
This work used the Extreme Science and Engineering Discovery Environment (XSEDE), which is supported by National Science Foundation grant number ACI-1548562. 
In particular, it used the Bridges-GPU AI system at the Pittsburgh Supercomputing Center (PSC) through allocations TG-CCR200004.

\bibliographystyle{acl_natbib}
\bibliography{emnlp2020, nlp}

\section{Appendix}

\subsection{Single-layer Sensitivity}

\begin{figure}[b]
\centering
\includegraphics[width=1.0\columnwidth]{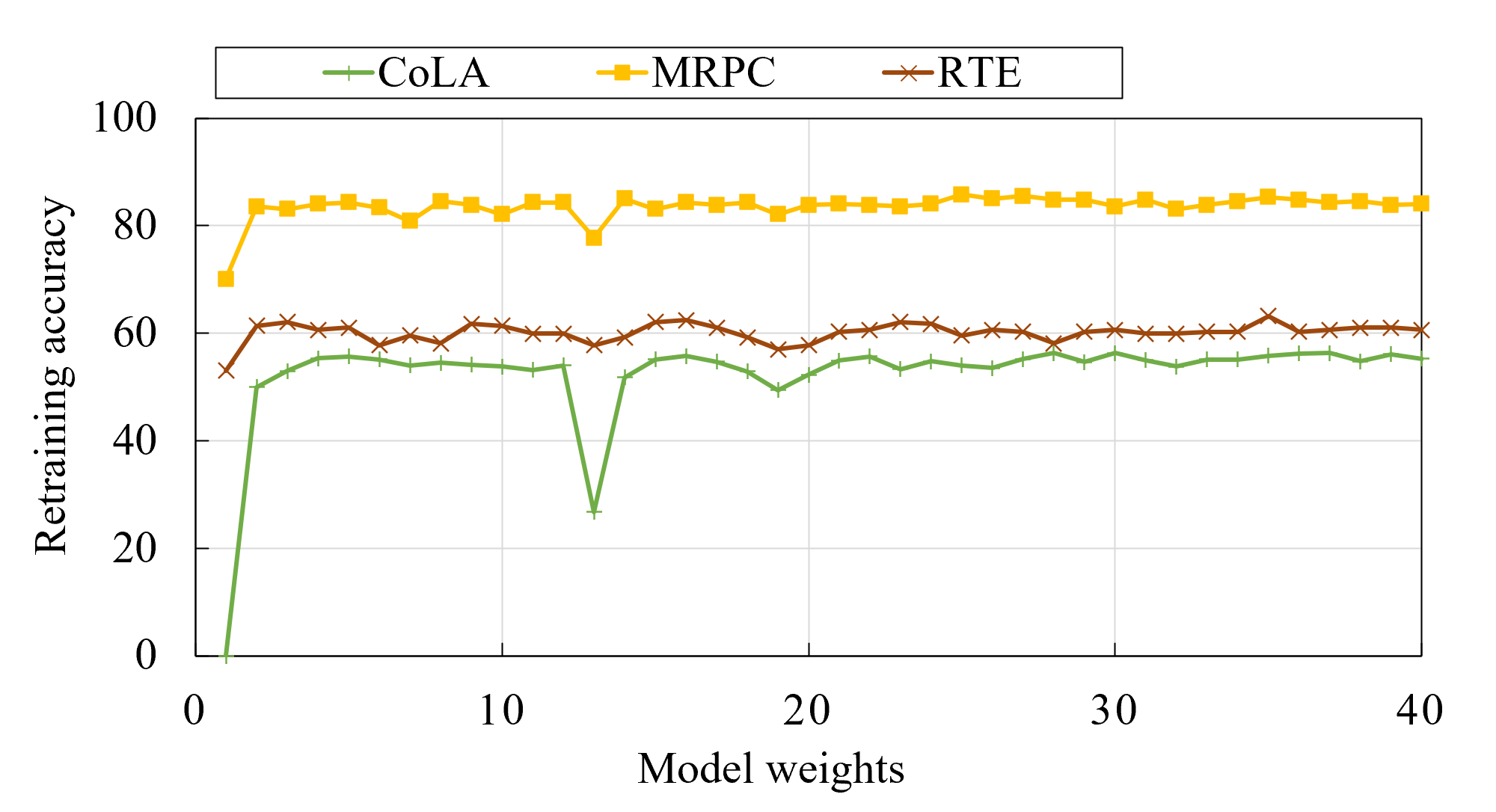}
\vskip -0.8em
\caption{Layer sensitivity with DistilBERT model.}
\label{fig:Layer_sensitivity}
\end{figure}

Before retraining, block structured pruning is carried out for the reweighted trained models by choosing compression ratio for each layer. However, the sensitivity of different layers are different, which may leads to significant accuracy loss if the compression ratios are not proper. To test the sensitivity, we prune 50\% of each layer while keeping the other layers unpruned and obtain the final accuracy after retraining. According to tests, embedding layers are sensitive on all datasets except WNLI. On MRPC and RTE datasets, we choose 8 as the number of blocks and $3e^{-4}$ as the penalty factor. In Figure~\ref{fig:Layer_sensitivity}, the first two weight matrices are related to embedding layers, while the third to the 38-th weights are related to transformer layers (each transformer layer includes 6 weights). The last two layers is related to classifier layers. The results show that the embedding layers and linear weights in transformer layers are sensitive on CoLA and MRPC datasets. Therefore, we set the compression ratios of corresponding weights zero to ensure the final accuracy.

\subsection{Number of Blocks}
\begin{figure}[b]
\centering
\includegraphics[width=1.0\columnwidth]{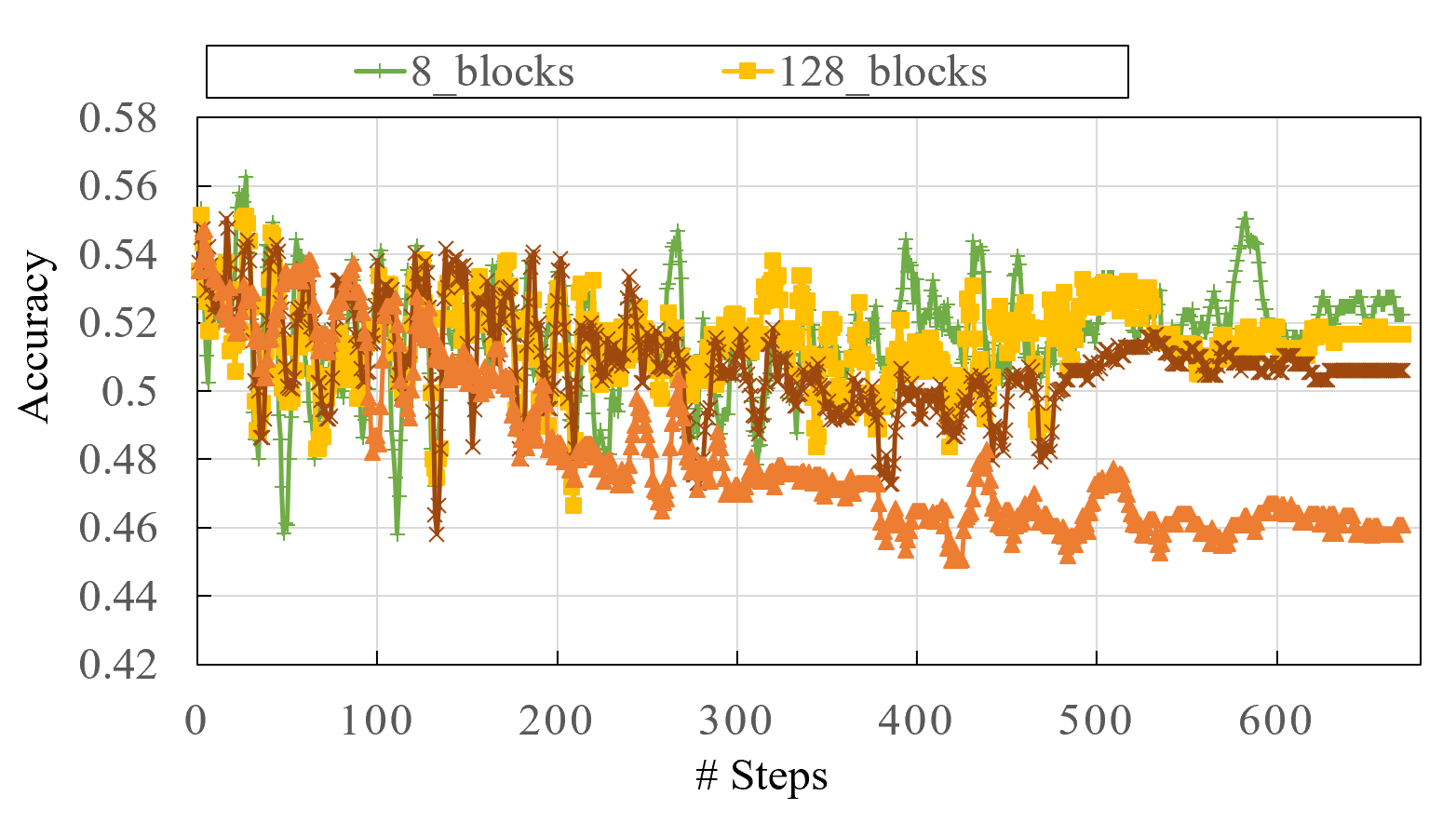}
\vskip -0.8em
\caption{Reweighted training accuracy of different weight matrix block division on CoLA dataset with DistilBERT model.}
\label{fig:Ablation_Num_blocks_RE_training_acc}
\end{figure}

\begin{figure}[]
\centering
\includegraphics[width=1.0\columnwidth]{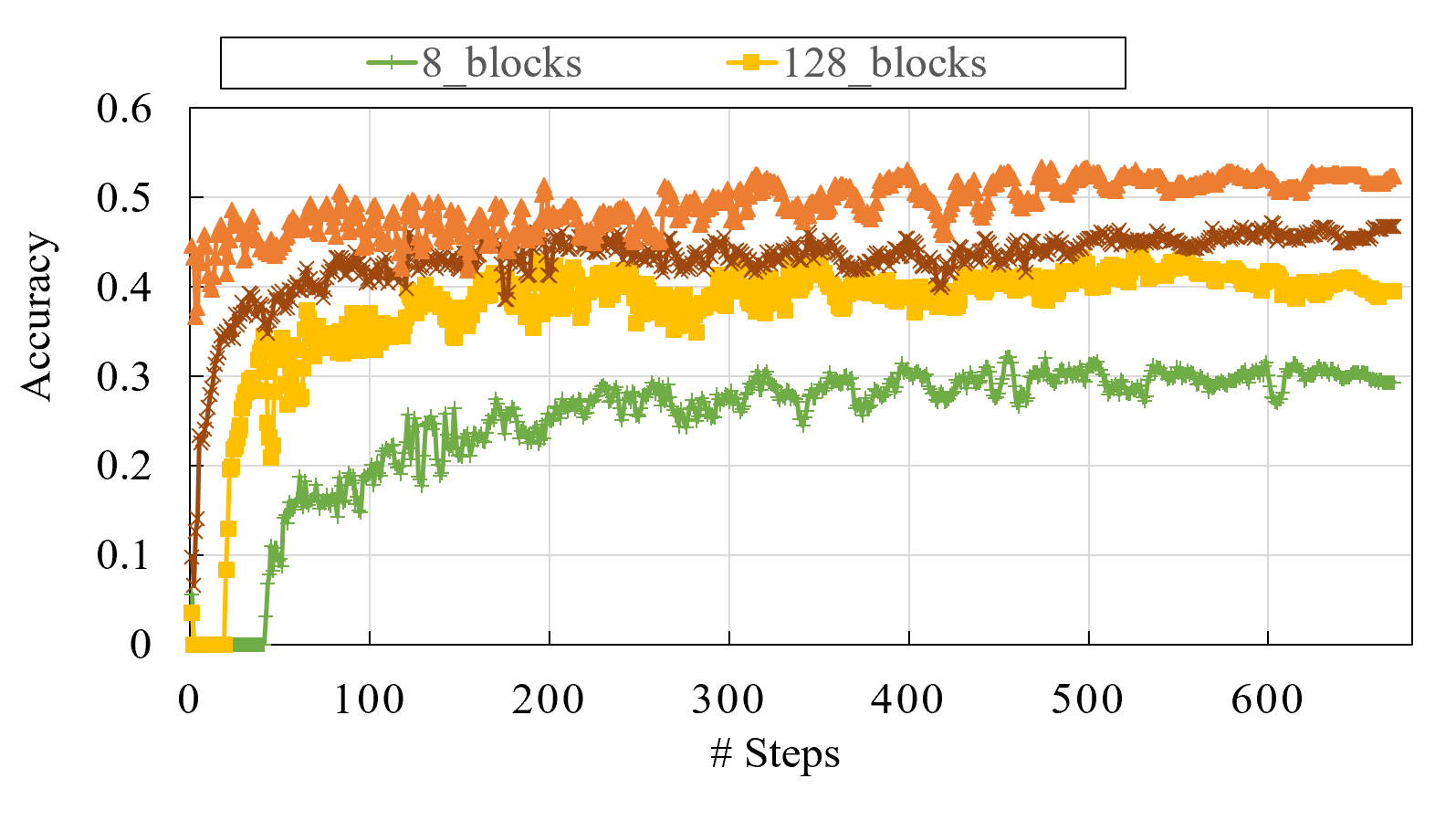}
\vskip -0.8em
\caption{Retraining accuracy of different weight matrix block division on CoLA dataset with DistilBERT model.}
\label{fig:Ablation_Num_blocks}
\end{figure}

\begin{figure}[]
\centering
\includegraphics[width=1.0\columnwidth]{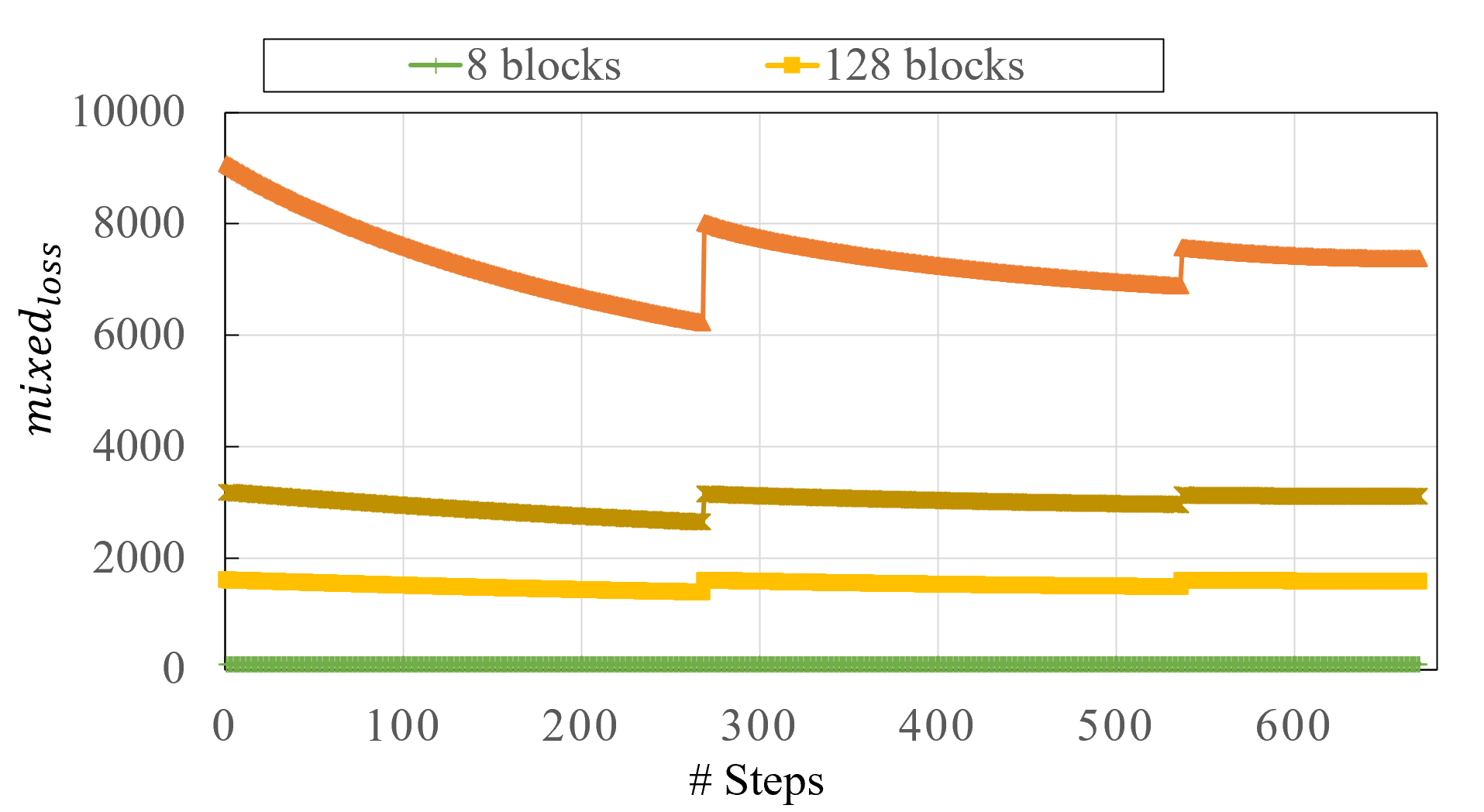}
\vskip -0.8em
\caption{Mixed loss during reweighted training of different weight matrix block divisions on CoLA dataset with DistilBERT model.}
\label{fig:Ablation_Num_blocks_mixed_loss}
\end{figure}

Figure~\ref{fig:Ablation_Num_blocks_RE_training_acc} and Figure~\ref{fig:Ablation_Num_blocks} represent reweighted training and retraining accuracy of different block sizes, respectively. During reweighted training, the accuracy decreases when we increase the number of blocks, since the corresponding $l_{1}$ loss increases significantly, which leads to $mixed_{loss}$ to increase as shown in Figure~\ref{fig:Ablation_Num_blocks_mixed_loss}. The final accuracy is improved when increasing the number of blocks since the algorithm is capable to operate on smaller units of the weight matrices. 

\begin{figure}[]
\centering
\includegraphics[width=1.0\columnwidth]{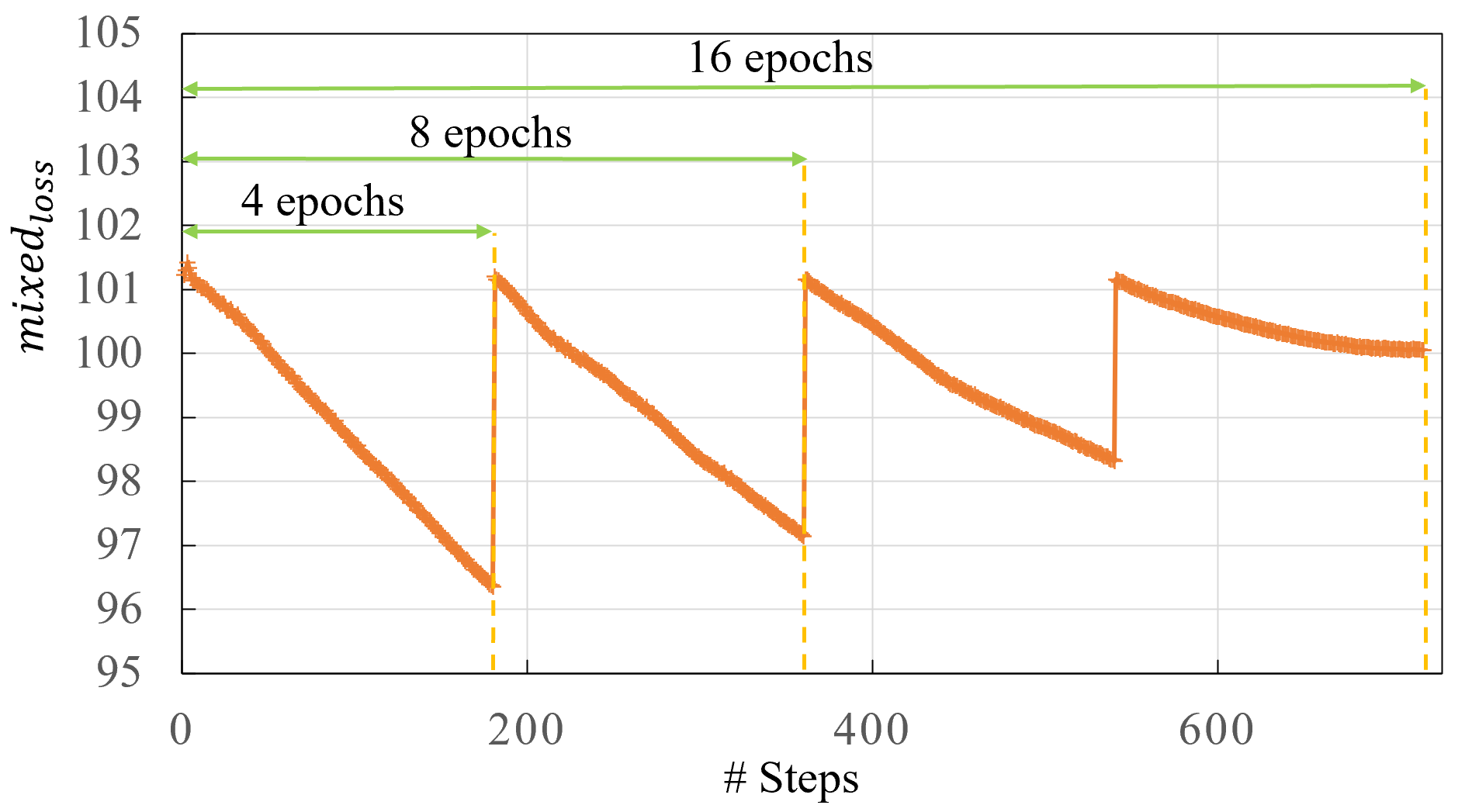}
\vskip -0.8em
\caption{Mixed loss of reweighted training with different epochs on STS-B dataset with DistilBERT model.}
\label{fig:Ablation_epochs_mixed_loss}
\end{figure}

\begin{figure}[]
\centering
\includegraphics[width=1.0\columnwidth]{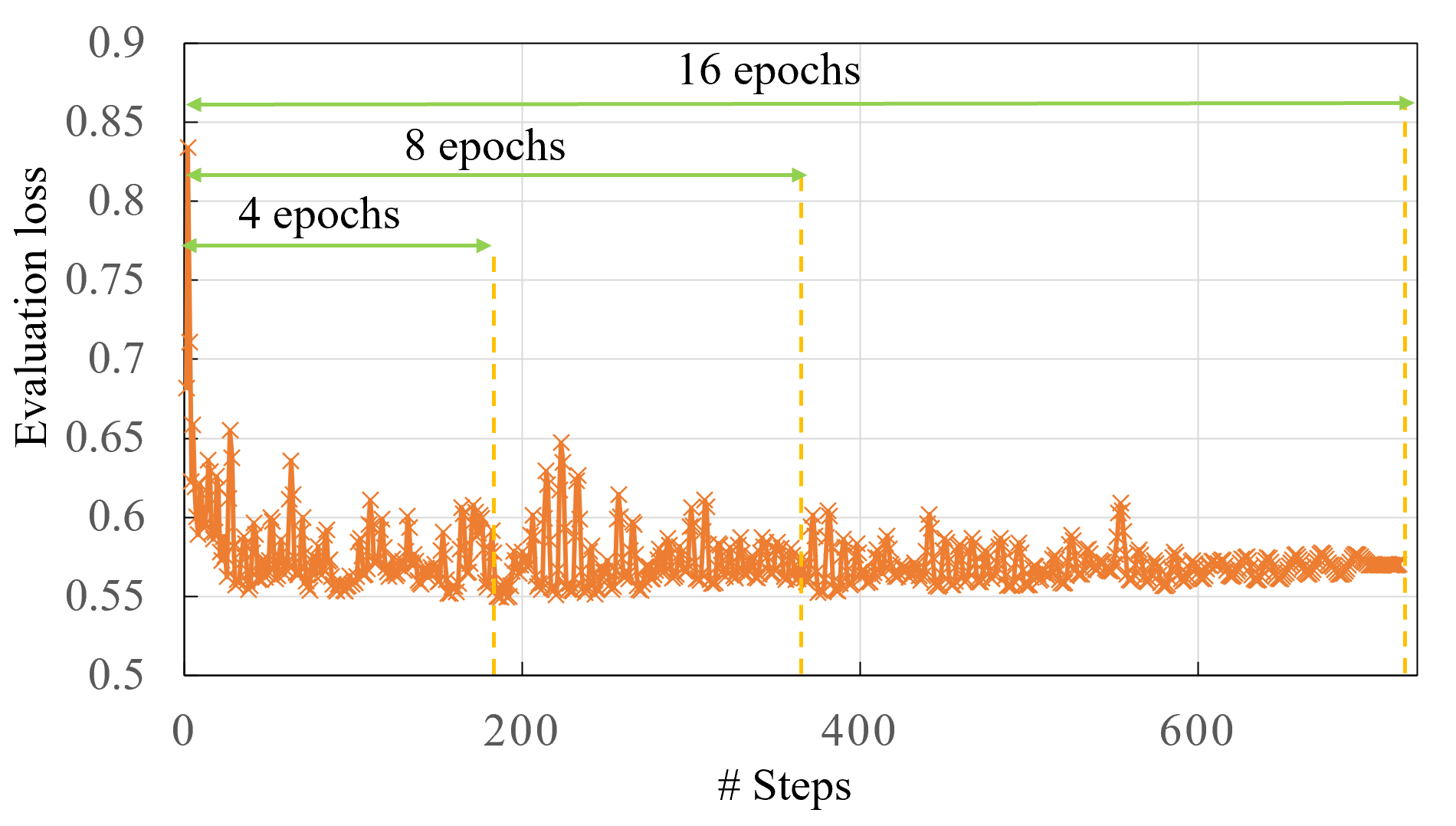}
\vskip -0.8em
\caption{Evaluation loss of reweighted training with different epochs on STS-B dataset with DistilBERT model.}
\label{fig:Ablation_epochs_eval_loss}
\end{figure}



\begin{figure}[]
\centering
\includegraphics[width=1.0\columnwidth]{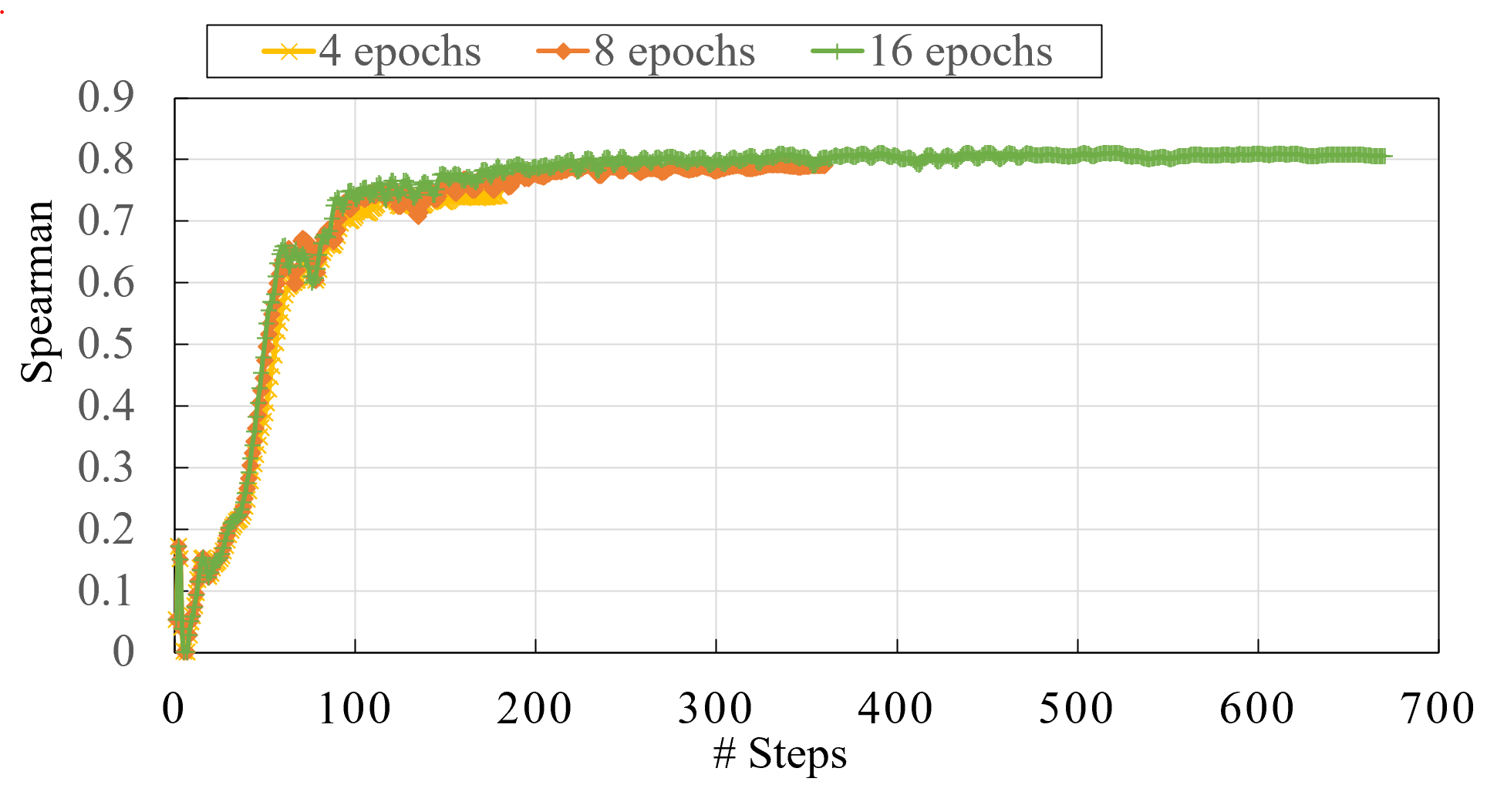}
\vskip -0.8em
\caption{Retraining spearman correlation with different retraining epochs on STS-B dataset with DistilBERT model.}
\label{fig:Ablation_epochs_retraining_spearman}
\end{figure}

\begin{figure}[]
\centering
\includegraphics[width=1.0\columnwidth]{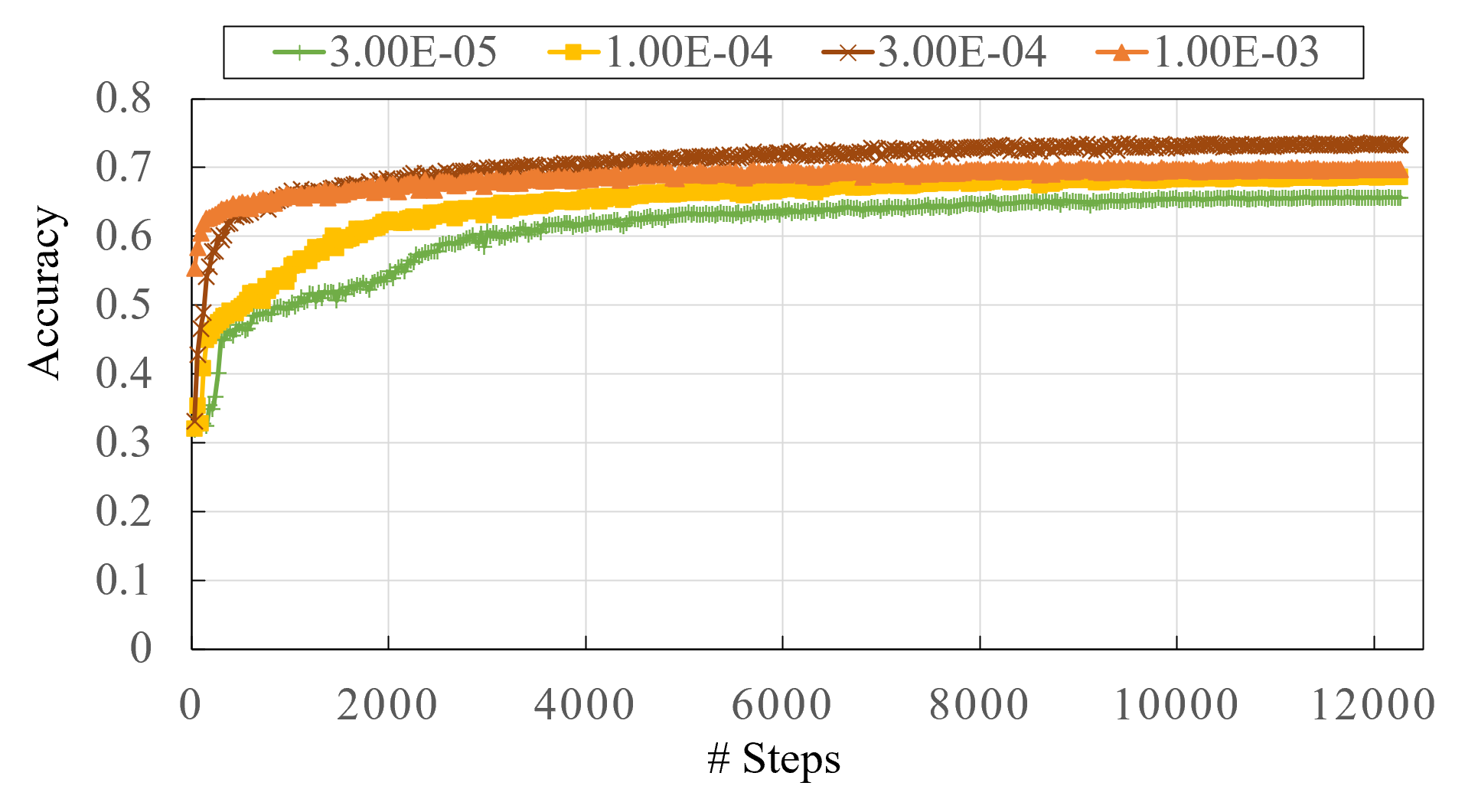}
\vskip -0.8em
\caption{Retraining accuracy using different penalty factors on MNLI dataset with DistilBERT model.}
\label{fig:Ablation_penalty}
\end{figure}

\subsection{Number of Retraining Epochs}
For reweighted training, Figure~\ref{fig:Ablation_epochs_mixed_loss} and Figure~\ref{fig:Ablation_epochs_eval_loss} show the results of mixed and evaluation loss, respectively, in which we update the $\gamma$ matrix every four epochs. 
For each selection of training epochs, we use linear learning rate decay and thus the results do not coincide with each other. The final accuracy of retraining is improved when we increase the training epochs as shown in Figure~\ref{fig:Ablation_epochs_retraining_spearman}. 



\subsection{Penalty Factors}
In Figure~\ref{fig:Ablation_penalty}, the retraining accuracy is improved when we increase the penalty factor from $3e^{-5}$ to $1e^{-4}$ and declines from $3e^{-4}$ to $1e^{-3}$.




\subsection{Retrain Accuracy}
\begin{figure}[h]
\centering
\includegraphics[width=0.9\columnwidth]{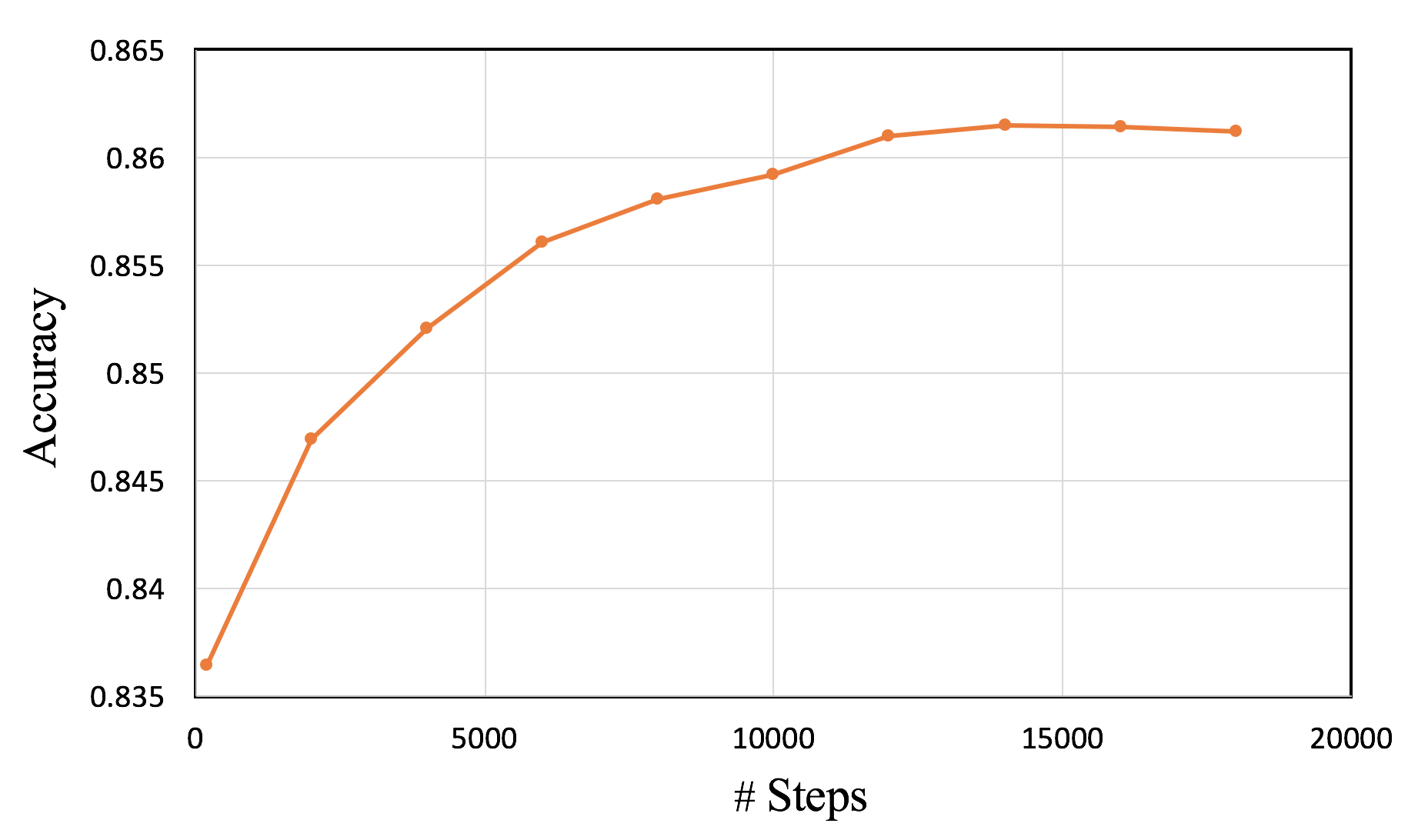}
\vskip -0.8em
\caption{Retraining accuracy on MNLI dataset with RoBERTa model.}
\label{fig:MNLI_retrain}
\end{figure}

\begin{figure}[]
\centering
\includegraphics[width=0.9\columnwidth]{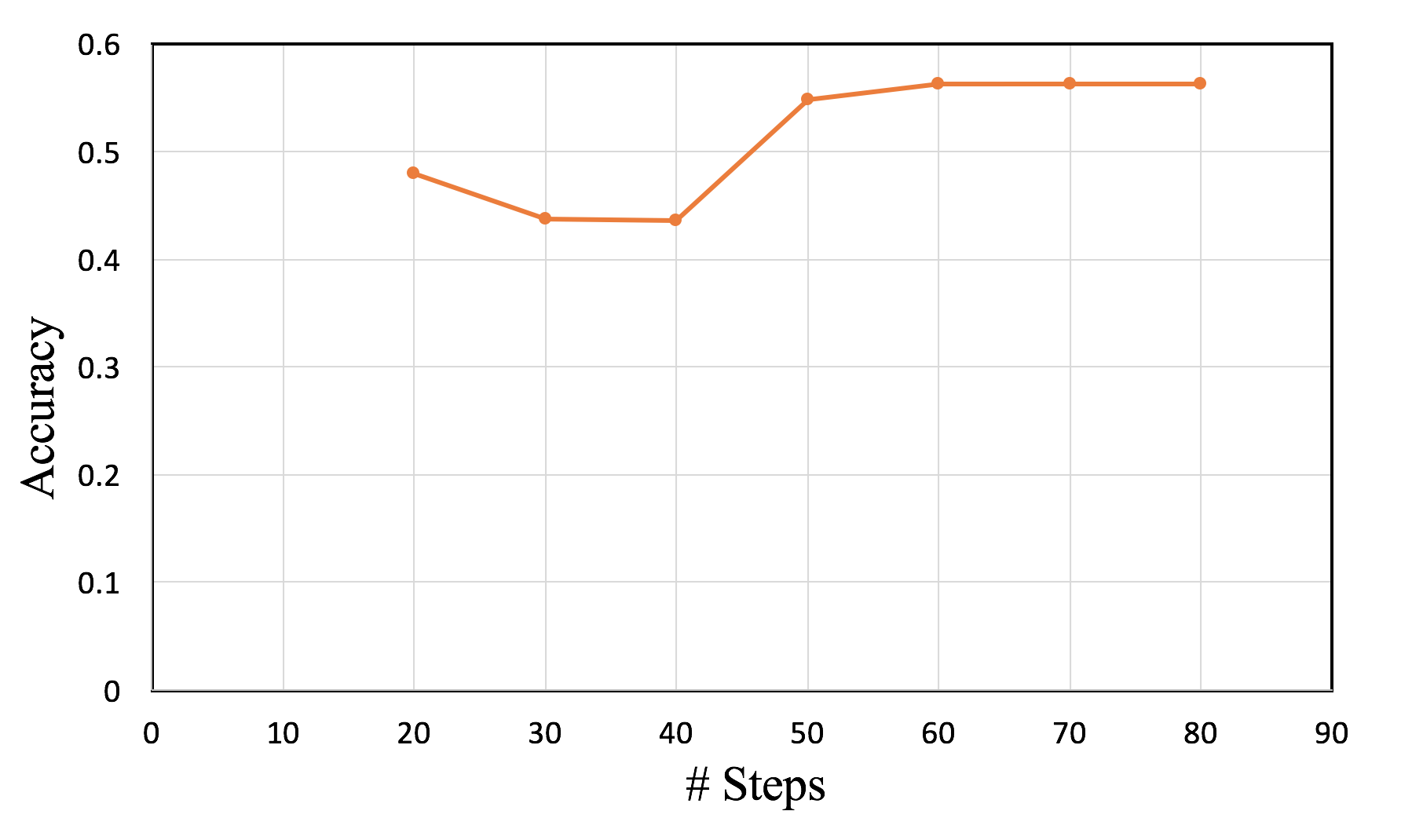}
\vskip -0.8em
\caption{Retraining accuracy on WNLI dataset with RoBERTa model.}
\label{fig:WNLI_retrain}
\end{figure}

Figure~\ref{fig:MNLI_retrain} $\sim$  Figure~\ref{fig:WNLI_retrain} show the accuracy with RoBERTa$_{\mathrm{BASE}}$ model on nine GLUE benchmark tasks during retraining steps.

\begin{figure}[]
\centering
\includegraphics[width=0.9\columnwidth]{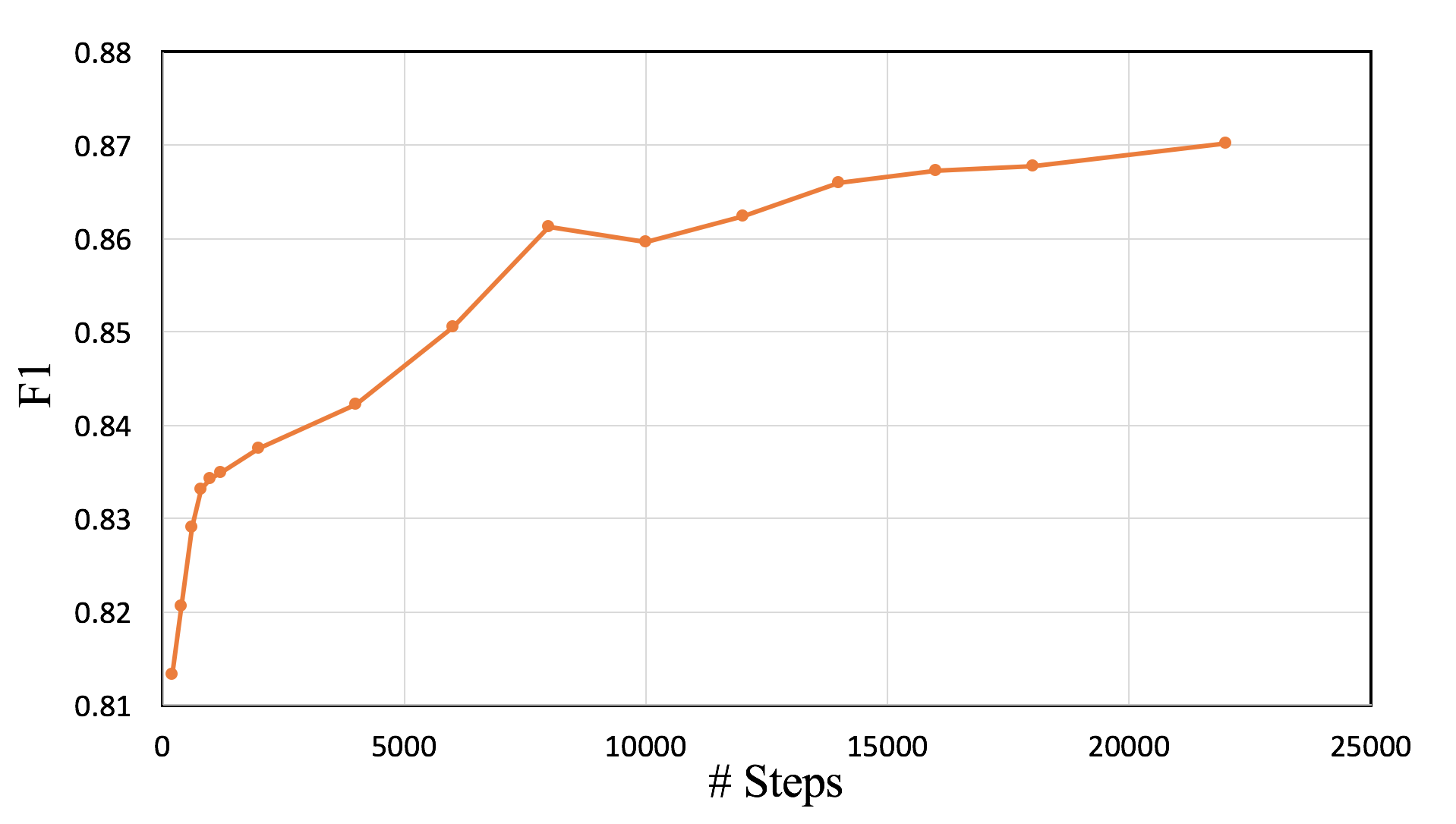}
\vskip -0.8em
\caption{Retraining F1 on QQP dataset with RoBERTa model.}
\label{fig:QQP_retrain}
\end{figure}

\begin{figure}[]
\centering
\includegraphics[width=0.9\columnwidth]{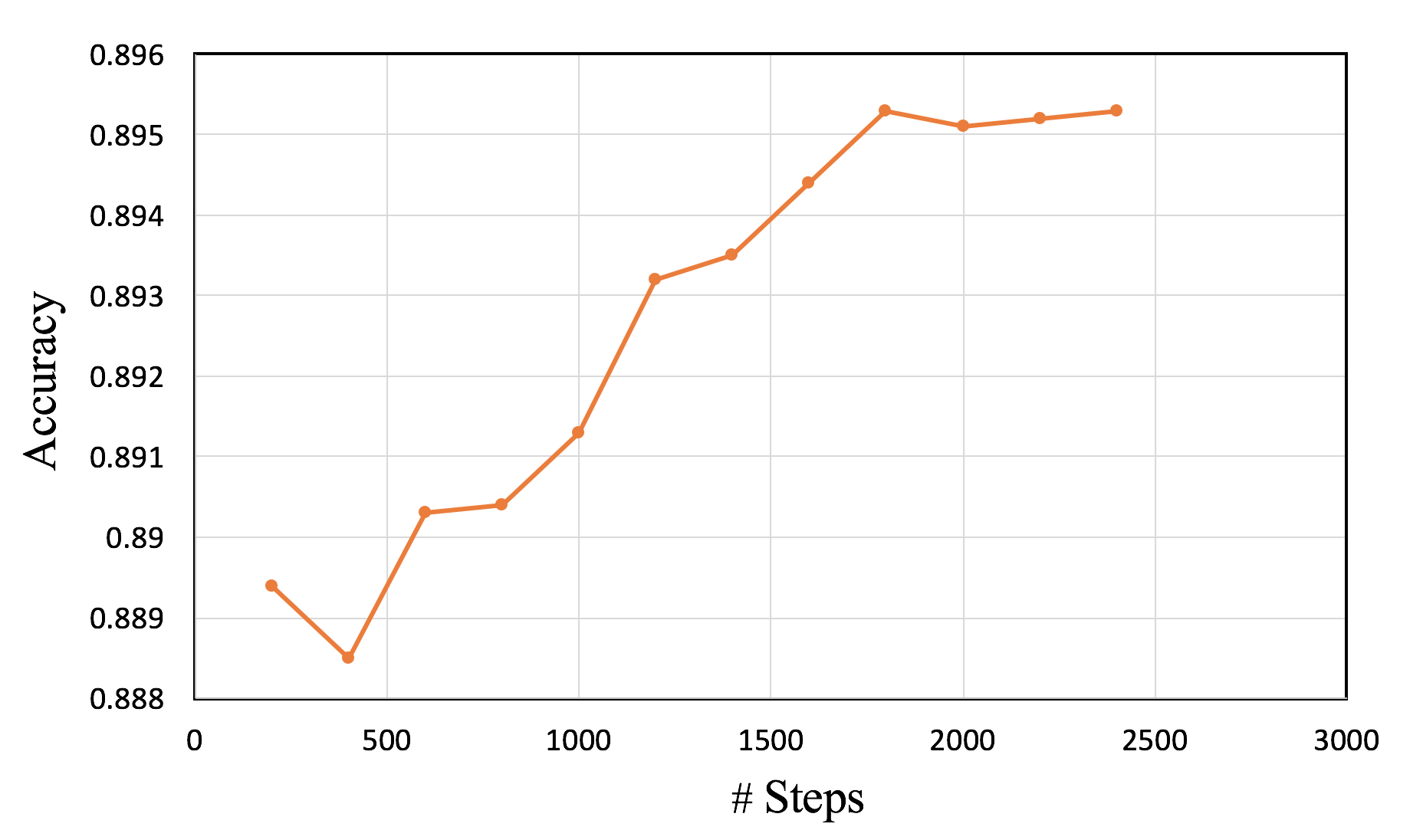}
\vskip -0.8em
\caption{Retraining accuracy on QNLI dataset with RoBERTa model.}
\label{fig:QNLI_retrain}
\end{figure}

\begin{figure}[]
\centering
\includegraphics[width=0.9\columnwidth]{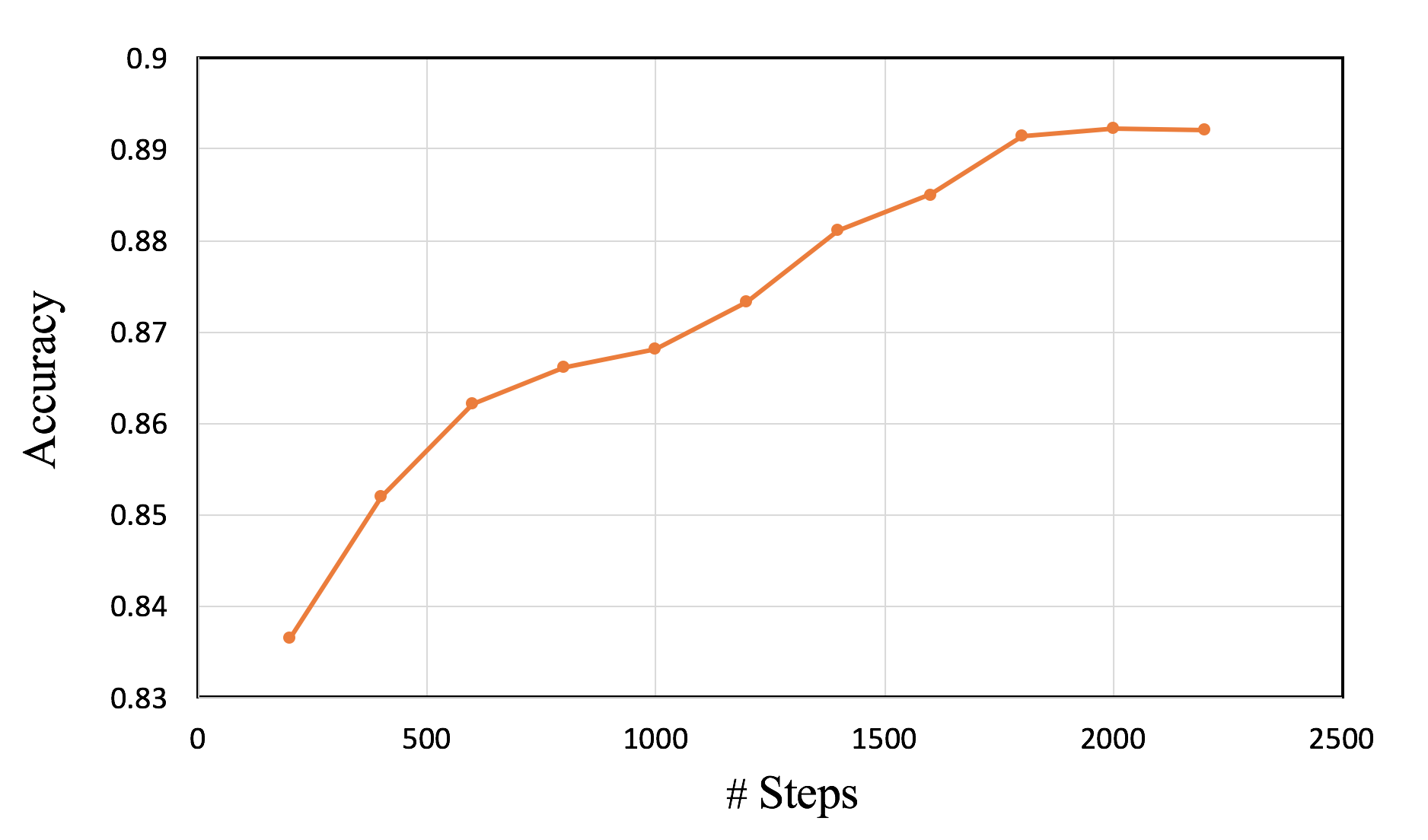}
\vskip -0.8em
\caption{Retraining accuracy on SST-2 dataset with RoBERTa model.}
\label{fig:SST-2_retrain}
\end{figure}

\begin{figure}[]
\centering
\includegraphics[width=0.9\columnwidth]{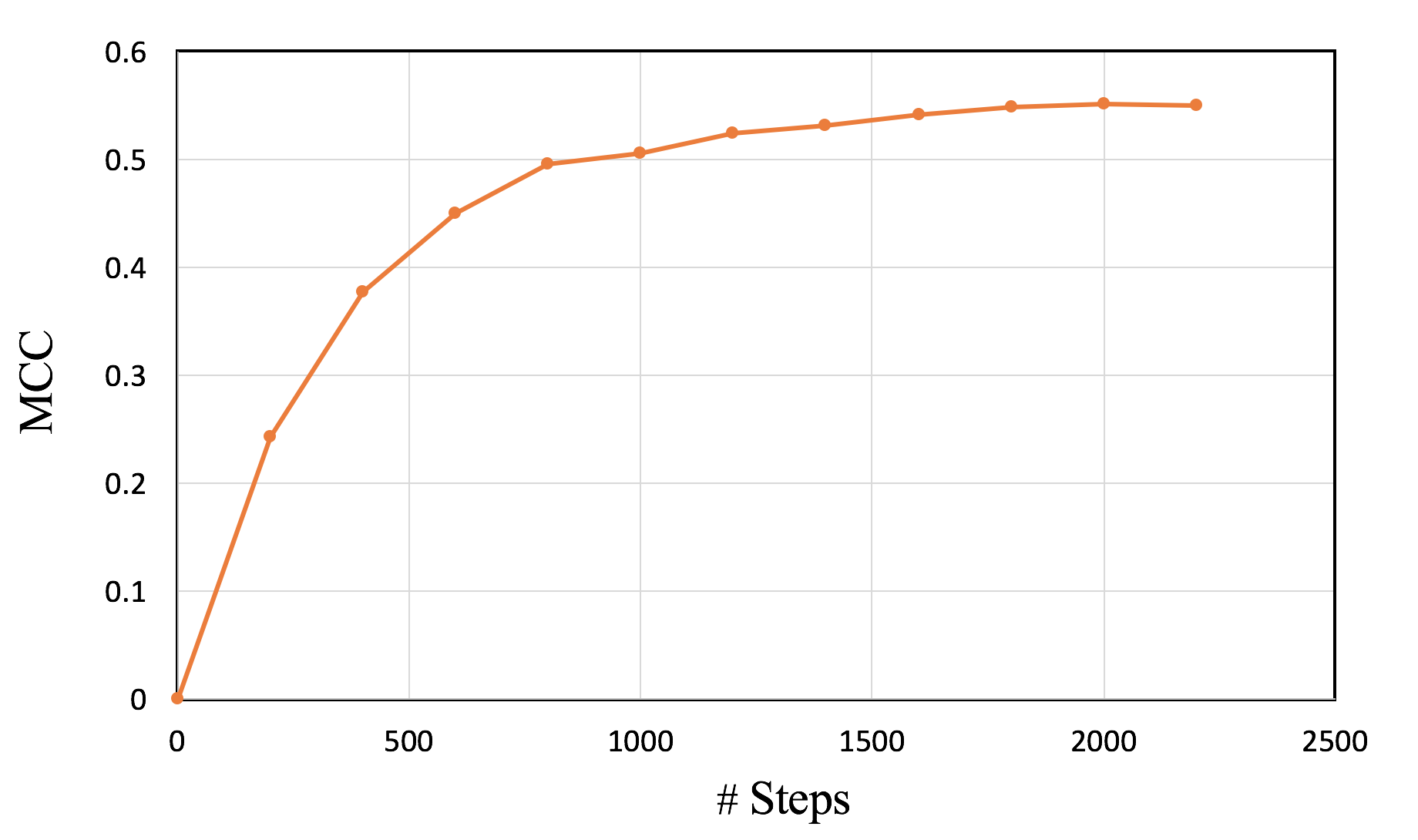}
\vskip -0.8em
\caption{Retraining mcc on CoLA dataset with RoBERTa model.}
\label{fig:CoLA_retrain}
\end{figure}

\begin{figure}[]
\centering
\includegraphics[width=0.9\columnwidth]{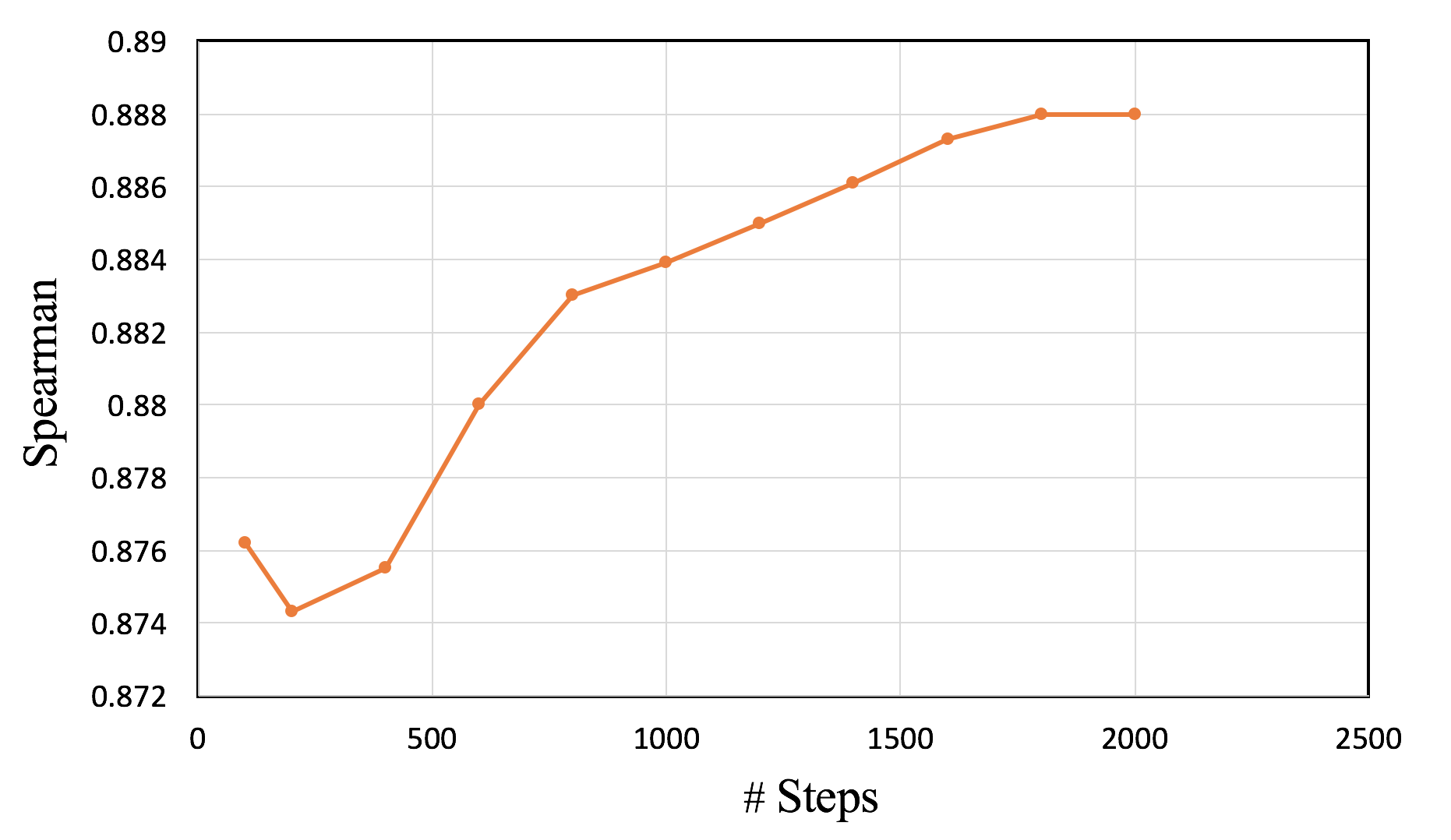}
\vskip -0.8em
\caption{Spearman correlation on STS-B dataset with RoBERTa model.}
\label{fig:STS-B_retrain}
\end{figure}

\begin{figure}[]
\centering
\includegraphics[width=0.9\columnwidth]{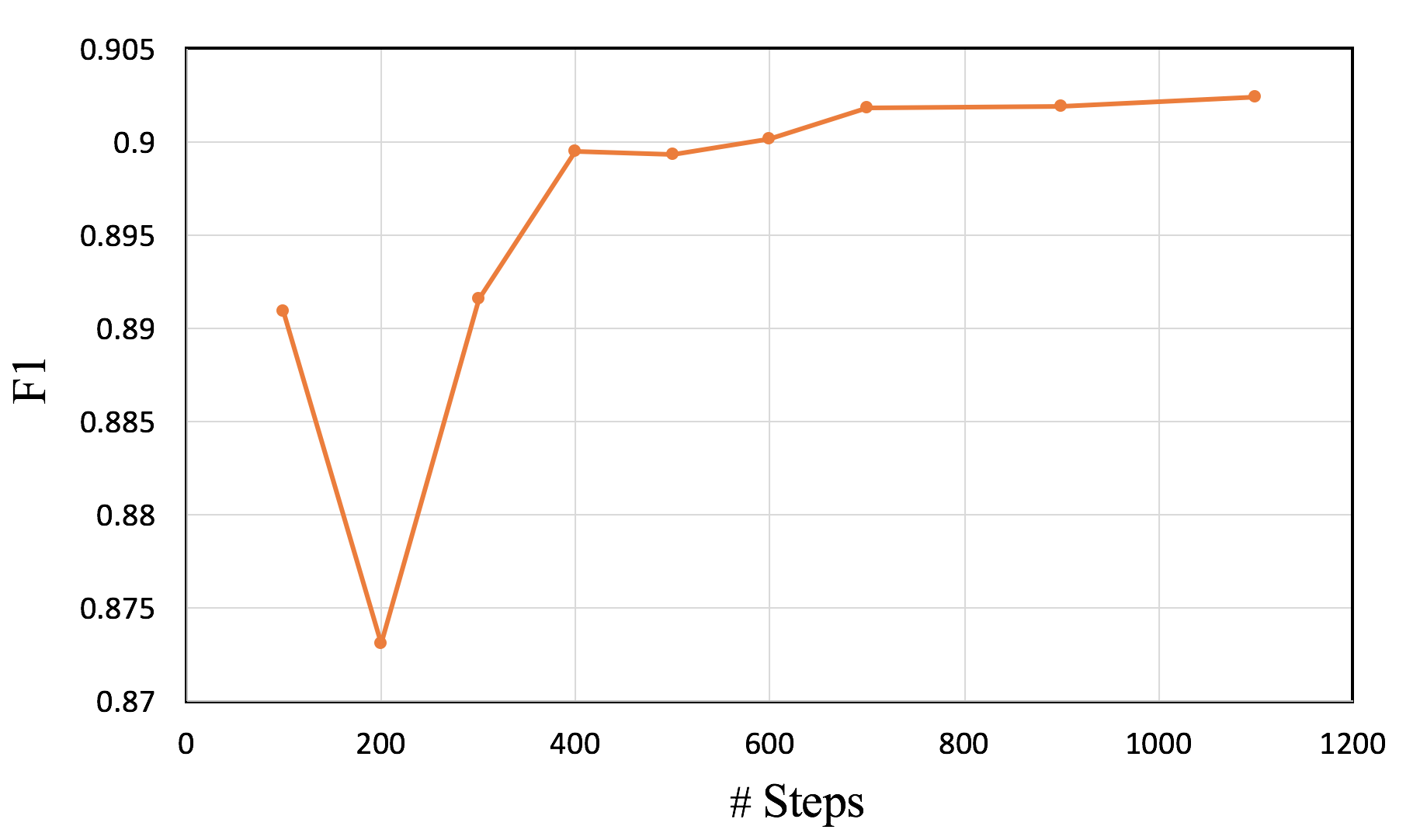}
\vskip -0.8em
\caption{Retraining F1 on MRPC dataset with RoBERTa model.}
\label{fig:MRPC_retrain}
\end{figure}

\begin{figure}[]
\centering
\includegraphics[width=0.9\columnwidth]{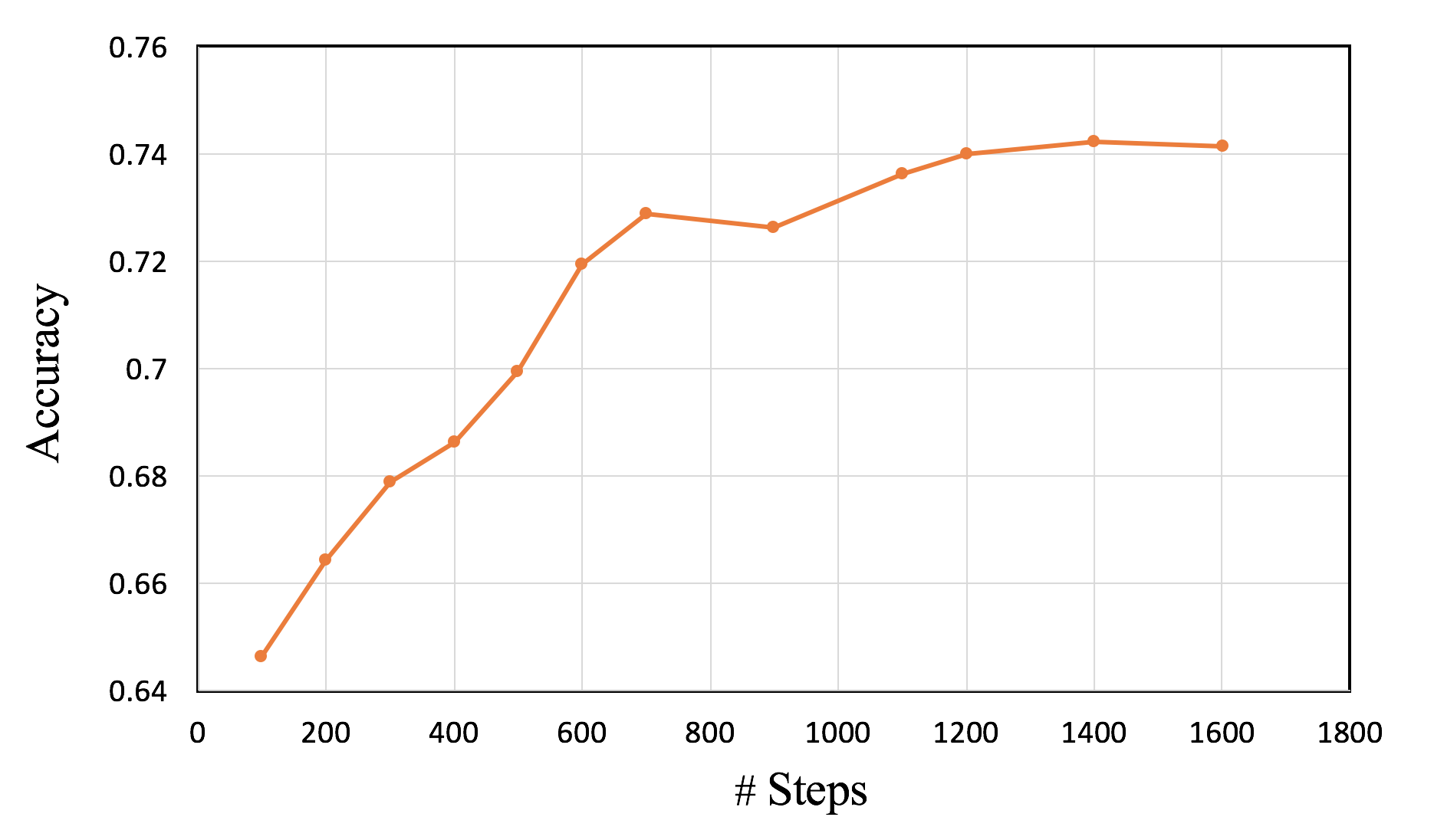}
\vskip -0.8em
\caption{Retraining accuracy on RTE dataset with RoBERTa model.}
\label{fig:RTE_retrain}
\end{figure}

\end{document}